\documentclass{article}
\usepackage{iclr2027_conference,times}

\iclrfinalcopy
\makeatletter
\let\cepo@original@maketitle\@maketitle
\def\@maketitle{\cepo@original@maketitle\lhead{Preprint}}
\makeatother

\usepackage[utf8]{inputenc} %
\usepackage[T1]{fontenc}    %
\usepackage{hyperref}       %
\usepackage{url}            %
\usepackage{booktabs}       %
\usepackage{amsfonts}       %
\usepackage{nicefrac}       %
\usepackage{microtype}      %
\usepackage[table,dvipsnames]{xcolor}
\usepackage{multirow}
\usepackage{colortbl}
\usepackage{diagbox}
\usepackage{subcaption}
\usepackage{wrapfig}
\usepackage{fvextra}
\usepackage{tabularx}
\usepackage{array}
\usepackage{makecell}
\usepackage{caption}
\usepackage{makecell}
\usepackage{caption}
\usepackage{xcolor}
\usepackage{soul}
\definecolor{coordpinkhl}{HTML}{FBE6E8}
\definecolor{execbluehl}{HTML}{E6F1FC}
\sethlcolor{yellow!20}
\newcommand{\coordhl}[1]{%
  {\sethlcolor{coordpinkhl}\hl{#1}}%
}
\newcommand{\exechl}[1]{%
  {\sethlcolor{execbluehl}\hl{#1}}%
}
\usepackage{xltabular}

\definecolor{coordbg}{HTML}{DDF3F8}
\definecolor{execbg}{HTML}{FBE7CF}

\definecolor{whatbg}{HTML}{DCEBFF}
\definecolor{howbg}{HTML}{FFF2B2}

\newcommand{\whattag}[1]{\colorbox{whatbg}{\strut\textbf{#1}}}
\newcommand{\howtag}[1]{\colorbox{howbg}{\strut\textbf{#1}}}

\newcommand{\whathl}[1]{\colorbox{whatbg}{\strut #1}}
\newcommand{\howhl}[1]{\colorbox{howbg}{\strut #1}}

\newcommand{\promptdots}{\textnormal{\ldots}}

\usepackage[table]{xcolor}

\definecolor{oursbg}{HTML}{E8F5E9}

\newcommand{\ourshl}[1]{%
  \begingroup
  \setlength{\fboxsep}{1pt}%
  \colorbox{oursbg}{\strut #1}%
  \endgroup
}

\newcommand{\promptcell}[1]{%
\begin{minipage}[t]{\linewidth}
\setlength{\fboxsep}{0.6pt}%
\fontfamily{cmtt}\selectfont\raggedright\sloppy
\emergencystretch=2em
\setlength{\parskip}{1.4pt}
#1
\end{minipage}
}

\usepackage[table]{xcolor}

\DefineVerbatimEnvironment{PromptVerbatim}{Verbatim}{
  fontsize=\small,
  breaklines=true,
  breakanywhere=true,
  frame=single,
  framesep=2mm
}

\title{CEPO: A General Framework for Improving Prompt Optimization in Multi-Agent Systems}

\author{%
  \begin{minipage}{\dimexpr\textwidth-2\tabcolsep\relax}
  \raggedright\normalfont
  {\small\textbf{Jongha Kim$^{1,*}$, Seohyun Lee$^{2,*}$, Jonathan Minho Chung$^{2}$, Jinsung Yoon$^{3}$, Hyunwoo J. Kim$^{2,\dagger}$}}\\[4pt]
  $^{1}$Korea University\quad $^{2}$KAIST\quad $^{3}$Google Cloud AI Research\\[2pt]
  {\small\texttt{jonghakim@korea.ac.kr, jinsungyoon@google.com}\\
  \texttt{\{seohyunlee,minho,hyunwoojkim\}@kaist.ac.kr}}
  \end{minipage}%
}
\hypersetup{
  hidelinks,
  pdftitle={CEPO: A General Framework for Improving Prompt Optimization in Multi-Agent Systems},
  pdfauthor={Jongha Kim, Seohyun Lee, Jonathan Minho Chung, Jinsung Yoon, Hyunwoo J. Kim}
}

\usepackage[table]{xcolor}
\usepackage{amsmath}
\usepackage{booktabs}
\usepackage{amssymb}
\usepackage{adjustbox}
\usepackage{algorithm}
\usepackage{algpseudocode}
\usepackage{amsmath}
\usepackage{multirow}
\DeclareMathOperator*{\argmax}{arg\,max}

\newcommand{\dq}[1]{``#1''}

\begin{document}

\maketitle
\begingroup
\renewcommand{\thefootnote}{\fnsymbol{footnote}}
\footnotetext[1]{Equal contribution.}
\footnotetext[2]{Corresponding author.}
\endgroup

\begin{abstract}
Multi-agent systems (MAS) require both effective coordination and specialized task execution. Existing prompt optimization approaches typically couple these functions within agent-specific prompts. We introduce Coordinator-Executor Prompt Optimization (CEPO), a general framework for improving prompt optimization in MAS. CEPO separates system-level coordination from agent-level execution through a shared coordinator prompt and agent-specific executor prompts, and guides their updates according to their respective roles. This structure allows shared coordination rules to be refined consistently across agents while keeping execution instructions separately adjustable. CEPO provides the prompt structure and optimization guidance, while existing prompt optimizers perform the underlying search. Extensive experiments spanning 11 benchmarks across four domains demonstrate that applying CEPO to four existing prompt optimizers consistently improves their performance. Ablations and qualitative analyses support the proposed design, while controlled studies on a document-understanding MAS further demonstrate improved optimization budget efficiency and performance scaling with the number of agents.
\end{abstract}

\section{Introduction}
Large language models (LLMs) have demonstrated strong capabilities in reasoning, tool use, and cross-modal grounding~\citep{achiam2023gpt,schick2023toolformer,huang2023language}, enabling multi-agent systems (MAS)~\citep{wu2024autogen,li2023camel,zhuge2024gptswarm} in which specialized agents collaborate on complex tasks. These systems have shown promising results across document understanding~\citep{han2025mdocagent,sun2025docagent}, mathematical reasoning~\citep{du2024improving,yuan2025reinforce}, code generation~\citep{islam2024mapcoder}, and video understanding~\citep{liu2026longvideoagent,wang2026think}. However, realizing their potential requires both effective system-level coordination and specialized agent-level execution. In other words, each agent must perform their assigned tasks well while working together toward the system's overall goal.

Prompt optimization automates the search for effective instructions and has expanded from single-agent settings~\citep{zhou2022large,yang2023large,pryzant2023automatic,ye2024prompt} to multi-agent systems~\citep{zhou2025multi,zhuge2024gptswarm}. Recent methods jointly optimize agent prompts or incorporate feedback from downstream interactions and system-level outcomes~\citep{han2025mapgdmultiagentpromptgradient,wang2026maspo}. However, even when these methods account for agent interactions, their prompt updates typically target only agent-specific prompts, within which system-level coordination and agent-level execution remain coupled. Revising shared coordination rules can therefore require consistent changes across multiple prompts with different execution responsibilities. This raises our central question: \dq{\textit{how can prompt structure and optimization guidance make existing prompt optimizers more effective for MAS?}}

We introduce Coordinator-Executor Prompt Optimization (CEPO), a general framework for improving prompt optimization in multi-agent systems. CEPO specifies how prompts are structured and how their updates are guided, while existing prompt optimizers carry out the underlying search. The framework combines two complementary components. First, coordinator-executor prompt decomposition introduces a coordinator prompt that is shared across all agents and defines system-level rules for how they coordinate. Each agent receives this shared coordinator prompt together with its own executor prompt, which specifies how it performs its assigned task. This structure allows a single coordination update to apply consistently across all agents while keeping each agent's execution instructions separately adjustable. Second, role-aligned optimization provides distinct meta-instructions that guide coordinator updates toward team-level collaboration and executor updates toward each agent's task execution. Together, these components make coordination and execution explicit optimization targets that can be refined using different underlying optimizers.

\begin{figure*}[t!]
    \centering
    \includegraphics[width=\textwidth]{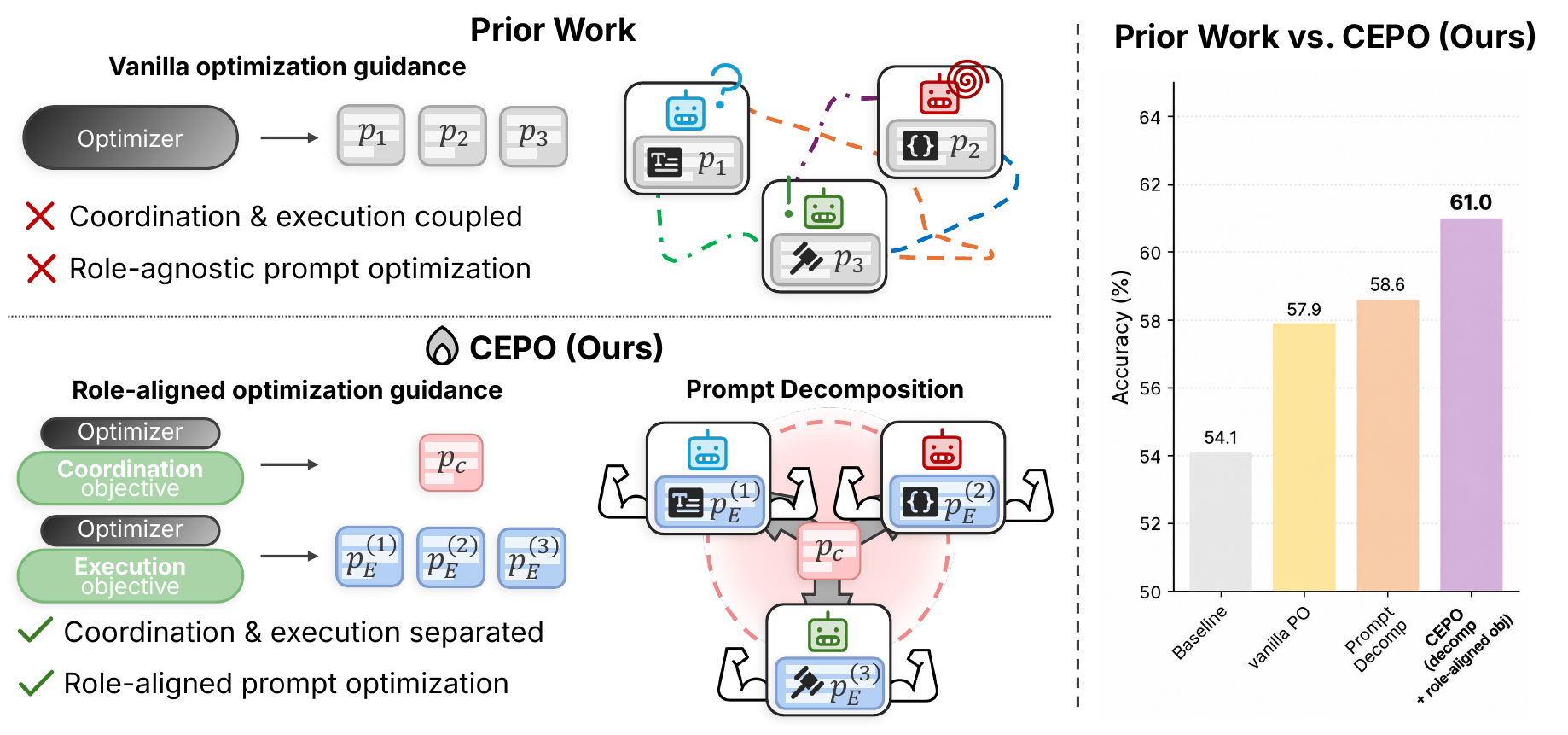}
    \vspace{-0.2cm}
    \caption{\textbf{CEPO separates coordination and execution for role-aligned prompt optimization.}
    \textbf{(a) Prior work vs. CEPO (left).}
    Existing methods couple coordination and execution roles within agent-specific prompts. CEPO represents these functions as distinct optimizable parameters through a shared coordinator prompt $p_C$ and agent-specific executor prompts $p_E$, and guides their updates according to their respective roles using an existing prompt optimizer.
    \textbf{(b) Component-wise comparison (right).} Accuracy on MDocAgent~\citep{han2025mdocagent} + MMLongBench-Doc~\citep{ma2024mmlongbench}. Vanilla optimizer improves baseline ($54.1 \to 57.9$), prompt decomposition yields a further gain ($58.6$), and combining decomposition with role-aligned optimization (CEPO) performs the best ($61.0$).}

\label{fig:motivation}
\end{figure*}
We apply CEPO to four different prompt optimizers (PE2, OPRO, ProTeGi, and MASPO)~\citep{ye2024prompt,yang2023large,pryzant2023automatic,wang2026maspo} and conduct extensive experiments on 11 benchmarks across four domains: document understanding, mathematical reasoning, code generation, and video understanding. Across the evaluated settings, CEPO consistently improves performance over the corresponding optimizers without CEPO, demonstrating its applicability to both general-purpose and MAS-specific prompt optimizers. Beyond these performance gains, controlled studies on MDocAgent show improved optimization budget efficiency and performance scaling with the number of agents. Component ablations, optimization-guidance swaps, and qualitative prompt analyses further support the benefits of separating coordination from execution and aligning optimization guidance with each component's role.

Our main \textbf{contributions} are as follows.

\begin{itemize}
\item We propose two principles to address the coupling of coordination and execution, a common limitation of existing MAS prompt optimization. We decompose prompts into a shared coordinator prompt for system-level coordination and agent-specific executor prompts for agent-level execution, and apply role-aligned guidance for each.

\item We integrate these two principles into CEPO, a general framework for MAS prompt optimization. Applying CEPO to four different prompt optimizers yields consistent gains in extensive experiments spanning eleven benchmarks across four domains.

\item Quantitative and qualitative analyses support the proposed design, and controlled studies on a document-understanding MAS further demonstrate improved optimization budget efficiency and performance scaling with MAS size.
\end{itemize}
\section{Related Work}
\label{sec:related}

\noindent\textbf{Multi-Agent Systems for Complex Reasoning.}
LLM-based multi-agent systems (MAS) distribute complex tasks across multiple agents. In document understanding, agents gather multimodal evidence, verify intermediate answers, and synthesize findings~\citep{han2025mdocagent,sun2025docagent,zhu2025doclens}. Mathematical reasoning systems use debate and iterative reflection to refine candidate solutions~\citep{du2024improving,yuan2025reinforce}. Code generation systems assign complementary roles for planning, programming, testing, and debugging~\citep{islam2024mapcoder,huang2023agentcoder}. Video understanding systems combine multimodal evidence processing, temporal localization, and hypothesis verification through agent collaboration~\citep{kugo2025videomultiagents,liu2026longvideoagent,wang2026think}. Across these domains, effective collaboration depends on both system-level coordination and agent-level execution, motivating the study of how prompts supporting these functions should be structured and optimized.

\noindent\textbf{Prompt Optimization.}
Prompt optimization automates the search for effective instructions, from single-agent tasks~\citep{zhou2022large,yang2023large,pryzant2023automatic,ye2024prompt,guo2023evoprompt} to compound systems~\citep{yuksekgonul2025optimizing,khattab2024dspy,opsahl2024optimizing}. Recent work extends optimization to MAS through joint search over prompts and topology~\citep{zhou2025multi,zhuge2024gptswarm} and collaborative optimization of agent prompts~\citep{han2025mapgdmultiagentpromptgradient,zhang2026mapro,hong2026maspob}. MASPO~\citep{wang2026maspo} further incorporates downstream interactions and coordination failures into prompt optimization. Although these approaches account for dependencies among agents, they typically do not explicitly separate system-level coordination and agent-level execution into distinct prompt components, nor provide role-aligned optimization guidance. To address this gap, we propose Coordinator-Executor Prompt Optimization (CEPO), a framework that introduces a shared coordinator prompt and agent-specific executor prompts and optimizes each according to its role, which complements the underlying prompt optimizer.
\section{Method}
\label{sec:method}
We propose Coordinator-Executor Prompt Optimization (CEPO), a general framework for improving existing prompt optimizers in multi-agent systems. CEPO combines two components: \emph{coordinator-executor prompt decomposition}, which separates shared coordination from agent-specific execution, and \emph{role-aligned optimization}, which guides updates to each component according to its function. We first formulate the MAS prompt optimization problem (\S\ref{sec:problem}), then introduce the two design principles and their implementation (\S\ref{sec:decomposition}), and finally describe the optimization procedure (\S\ref{sec:algorithm}).

\subsection{Problem Formulation}
\label{sec:problem}

We consider a multi-agent system $f_\mathcal{S}$ composed of $N$ LLM-based agents $\{A_1,\ldots,A_N\}$, each parameterized by a natural-language prompt $p_i$. Given a task query $q$ and optional context $x$, the system produces an output $\hat{y}=f_\mathcal{S}(x,q;\,\mathbf{p})$, where $\mathbf{p}=(p_1,\ldots,p_N)$. The context may contain task-specific information. Let $\mathcal{M}$ denote a task-specific performance metric and $\mathcal{D}_{\text{opt}}=\mathcal{D}_{\text{train}}\cup\mathcal{D}_{\text{val}}$ denote disjoint training and validation sets. Each example $(x,q,y)$ includes a reference $y$ used to evaluate the system output. The prompt optimization objective is:

\begin{equation}
\label{eq:objective}
    \mathbf{p}^{\star} \;=\; 
    \argmax_{\mathbf{p}}\; 
    \mathbb{E}_{(x, q, y) \sim \mathcal{D}_{\text{val}}}
    \big[\, \mathcal{M}\!\left(f_{\mathcal{S}}(x, q;\, \mathbf{p}),\, y\right) \,\big].
\end{equation}

Following common practice in prompt optimization~\citep{pryzant2023automatic,ye2024prompt}, training examples provide feedback for prompt updates, and validation performance selects prompts along the optimization trajectory. The feedback and selection mechanisms used within prompt search depend on the underlying optimizer.

Existing prompt optimizers~\citep{zhou2022large,yang2023large,pryzant2023automatic,ye2024prompt,zhou2025multi} can account for dependencies among agents when updating the prompt collection $\mathbf{p}$ in Eq.~\eqref{eq:objective}. Nevertheless, shared coordination and agent-specific execution remain expressed within per-agent prompts. We instead make them distinct optimization targets: a shared component specifies how agents collaborate, and local components specify how they execute their respective tasks. This separation allows a coordination update to be applied consistently across agents while execution updates remain specific to individual roles.

\subsection{Two Design Principles for MAS Prompt Optimization}
\label{sec:decomposition}

\noindent\textbf{Principle 1: Coordinator-executor prompt decomposition.}
To separate the role of system-level coordination and agent-level execution, we represent system-level coordination through a dedicated, shared prompt parameter. We implement this principle through \emph{coordinator-executor prompt decomposition}, a structured parameterization with a shared coordinator prompt $p_C$ and agent-specific executor prompts $\{p_E^{(1)},\ldots,p_E^{(N)}\}$. The coordinator specifies system-level collaboration, such as how agents share findings, flag disagreements, and support subsequent stages. Each executor specifies how its agent carries out its assigned task. At each invocation of agent $A_i$, the effective prompt is

\begin{equation}
\label{eq:concat}
    \tilde{p}_i \;=\; p_C \oplus p_E^{(i)},
\end{equation}

where $\oplus$ denotes prepending $p_C$ to $p_E^{(i)}$. The full parameterization becomes $\mathbf{P}=\big(p_C,p_E^{(1)},\ldots,p_E^{(N)}\big)$, and Eq.~\eqref{eq:objective} is reformulated as $\mathbf{P}^{\star}=\argmax_{\mathbf{P}} J(\mathbf{P})$, with $J(\mathbf{P})$ denoting the validation objective in Eq.~\eqref{eq:objective} under the substitution $p_i\mapsto\tilde{p}_i$. Updating $p_C$ changes the shared instructions received by all agents, whereas updating $p_E^{(i)}$ changes the local instructions of agent $A_i$.

\noindent\textbf{Principle 2: Role-aligned optimization.}
To ensure that the coordinator and executor prompts are optimized for their intended roles, we propose role-aligned optimization. We implement this strategy using role-specific optimization meta-instructions, $\Phi_C$ and $\Phi_E$, which are supplied to the underlying optimizer.
When updating $p_C$, $\Phi_C$ directs the optimizer to refine system-level collaboration, including cross-agent communication, handling of conflicting evidence, and shared reasoning norms. When updating $p_E^{(i)}$, $\Phi_E$ directs it to refine the target agent's execution, including task scoping, evidence use, and concrete procedures under the shared coordination context. Thus, $\Phi_C$ provides system-level guidance on how agents should coordinate, whereas $\Phi_E$ provides agent-level guidance on how each agent should execute its assigned task.

Note that both prompt layers are selected according to the system-level performance objective $J$. The meta-instructions provide different guidance for their updates, although they do not define separate numerical objectives. The meta-instruction excerpts in Table~\ref{tab:mdocagent_meta_instruction_excerpt} illustrate this distinction, and the swap analysis in Table~\ref{tab:swap} evaluates the effect of aligning the guidance with each prompt's role.

\begin{table*}[t]
\centering
\caption{\textbf{Meta-instructions for role-aligned guidance.} 
The coordinator meta-instruction $\Phi_C$ directs the optimizer toward \coordhl{team-level coordination signals}, while the executor meta-instruction $\Phi_E$ directs it toward \exechl{agent-level execution signals}.
Full prompts reported in Appendix~\ref{sec:optimizer_llm_prompt} and~\ref{sec:agent_optimizer_llm_prompt}.}
\vspace{-0.2cm}
\label{tab:mdocagent_meta_instruction_excerpt}
\scriptsize
\setlength{\tabcolsep}{5pt}
\renewcommand{\arraystretch}{1.08}
\begin{tabularx}{\textwidth}{
@{}>{\raggedright\arraybackslash}X
>{\raggedright\arraybackslash}X@{}}
\toprule
\textbf{Coordinator optimizer meta-instruction ($\Phi_C$)} &
\textbf{Executor optimizer meta-instruction ($\Phi_E$)} \\
\midrule

{\fontfamily{cmtt}\selectfont
You are optimizing one \coordhl{shared multi-agent system prompt} for a multi-agent document QA pipeline. This shared prompt is prepended to ALL five agents. Each agent also receives its own agent-specific prompt with role-specific instructions. 

The goal is to improve \coordhl{system-level behavior across the trace.} \textnormal{\quad$\cdots$\quad} Prefer \coordhl{compact shared rules that help the pipeline make better decisions} over formatting instructions that only change how answers look.
}
&
{\fontfamily{cmtt}\selectfont
Your task is to optimize only the \exechl{target agent's executor prompt}: the how-to instructions that tell this specific agent how to perform \textnormal{\quad$\cdots$\quad}Infer the execution behavior that would have prevented it and convert it into reusable executor-level instructions. 
The final prompt should be \exechl{concrete and operational} \textnormal{\quad$\cdots$\quad}  
\exechl{provide enough task-specific guidance} to avoid recurring failure patterns. 
\exechl{Include procedural details} when they clarify how the agent should select evidence.
}
\\
\bottomrule
\end{tabularx}
\vspace{-0.4cm}
\end{table*}
\noindent\textbf{Coordinator-Executor Prompt Optimization.}
CEPO combines coordinator-executor prompt decomposition with role-aligned optimization. The framework specifies the prompt components and the guidance for optimizing them, while an underlying prompt optimizer performs the search and updates. For the single-target update procedure below, we write an optimizer call as $p'=\mathcal{O}(p;\,\mathcal{C},\,\Phi,\,\mathcal{D}_{\text{opt}})$, where $p$ is the target prompt, $\mathcal{C}$ contains the other prompts held fixed, $\Phi$ is the role-specific optimization meta-instruction, and $\mathcal{D}_{\text{opt}}$ supplies the optimization data. Candidate generation and search follow the underlying optimizer, and we show that  CEPO is generally applicable to various optimizers, such as PE2~\citep{ye2024prompt}, OPRO~\citep{yang2023large}, ProTeGi~\citep{pryzant2023automatic} (Table~\ref{tab:optimization_algorithm}), and MASPO~\citep{wang2026maspo} (Table~\ref{tab:maspo}).

\begin{figure*}[t!]
    \centering
    \includegraphics[width=\textwidth]{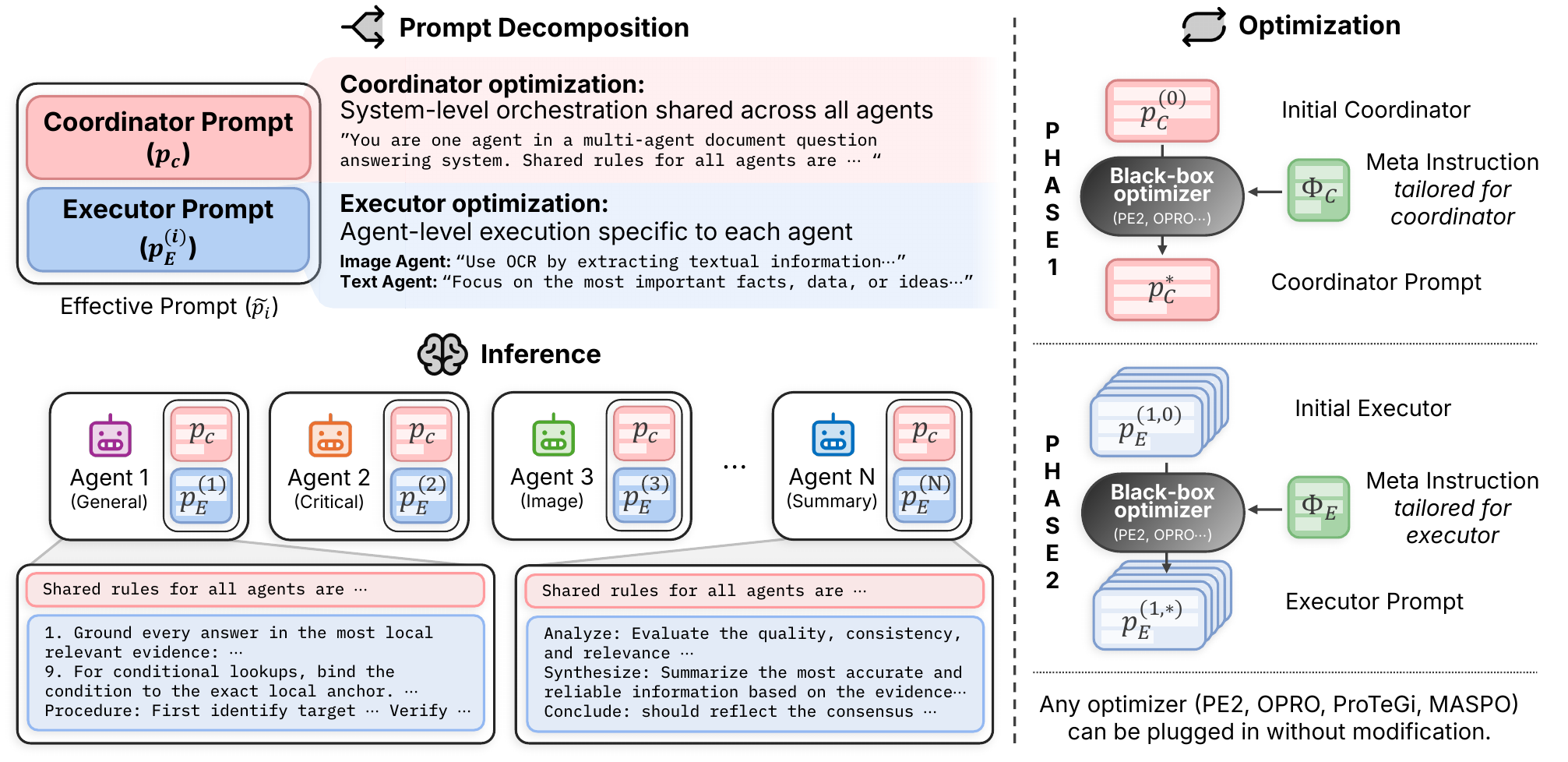}
    \vspace{-0.4cm}
    
    \caption{\textbf{Overview of CEPO.} CEPO decomposes prompts within a multi-agent system into two layers with distinct roles. The \emph{coordinator prompt} $p_C$ (top), shared across all agents, captures coordination signals (team composition, role assignment, and shared norms for reasoning). Each \emph{executor prompt} $p_E^{(i)}$ (bottom), local to agent $A_i$, captures execution signals (evidence anchoring, task scoping, and procedure execution). The effective prompt for $A_i$ is the concatenation $p_C \oplus p_E^{(i)}$. Two layers are sequentially optimized with role-aligned guidance under CEPO: the coordinator is optimized using meta-instruction $\Phi_C$ with executors fixed (Phase~1), then executors are optimized using $\Phi_E$ with the coordinator fixed (Phase~2). CEPO applies these design principles to existing prompt optimizers $\mathcal{O}$, consistently improving those methods.}
\label{fig:overview}
\end{figure*}
\subsection{The CEPO Optimization Procedure}
\label{sec:algorithm}

Our default procedure first optimizes the coordinator with the initial executor prompts fixed, then optimizes selected executors under the resulting shared coordination context. This schedule performs best among those evaluated in our document-system ablation (Appendix~\S\ref{sec:supp_optimization_schedule}). It is an implementation choice, while CEPO is defined by prompt decomposition and role-aligned optimization.

\noindent\textbf{Phase 1: Coordinator optimization.}
Starting from an initial coordinator $p_C^{(0)}$ and fixed initial executors $\{p_E^{(i,0)}\}$, we update the coordinator for $T_C$ iterations using $\Phi_C$:

\begin{equation}
\label{eq:coord-update}
    p_C^{(t)} \;=\; \mathcal{O}\!\left(p_C^{(t-1)};\, \{p_E^{(i,0)}\}_{i=1}^{N},\, \Phi_C,\, \mathcal{D}_{\text{opt}}\right), 
\quad t = 1, \ldots, T_C.
\end{equation}

Here, $\mathcal{O}(p;\,\mathcal{C},\,\Phi,\,\mathcal{D})$ updates the target prompt while keeping $\mathcal{C}$ fixed. The full MAS $f_\mathcal{S}$ is run on $\mathcal{D}_{\text{train}}$, and failures under $\mathcal{M}$ provide feedback for candidate updates. Each coordinator iterate is evaluated on $\mathcal{D}_{\text{val}}$ with the initial executors fixed. We select $p_C^{\star}$ from the initial prompt and all iterates according to the highest system-level validation score $J$.

\noindent\textbf{Phase 2: Executor optimization.}
With $p_C^{\star}$ fixed, we optimize the executor prompts of a user-specified subset $\mathcal{I}\subseteq\{1,\ldots,N\}$. Each selected executor is updated for $T_E$ iterations using $\Phi_E$:

\begin{equation}
\label{eq:exec-update}
    p_E^{(i,t)} \;=\; \mathcal{O}\!\left(p_E^{(i,t-1)};\, p_C^{\star},\, \{p_E^{(j,0)}\}_{j \neq i},\, \Phi_E,\, \mathcal{D}_{\text{opt}}\right),
\quad t = 1, \ldots, T_E.
\end{equation}

Training feedback is collected from the full MAS under $p_C^{\star}$ and the executor context used for the update, and the executor prompt is selected by validation scores. Agents outside $\mathcal{I}$ retain their initial prompts. The choice $\mathcal{I}=\emptyset$ gives coordinator-only optimization, while $\mathcal{I}=\{1,\ldots,N\}$ permits optimization of every executor. The executor subsets are described in \S\ref{sec:supp_optimization_agent_set}. Algorithm~\ref{alg:cepo} summarizes the default procedure.

\begin{algorithm}[t]
\caption{CEPO optimization procedure.}
\label{alg:cepo}
\begin{algorithmic}[1]
\Require MAS $f_\mathcal{S}$ with $N$ agents; training set 
    $\mathcal{D}_{\text{train}}$ and validation set 
    $\mathcal{D}_{\text{val}}$; metric $\mathcal{M}$; 
    optimizer $\mathcal{O}$; meta-instructions $\Phi_C, \Phi_E$; 
    initial prompts $p_C^{(0)}, \{p_E^{(i,0)}\}_{i=1}^{N}$; 
    iteration budgets $T_C, T_E$; agent subset $\mathcal{I}$.
\Ensure Optimized prompt set 
    $\mathbf{P}^{\star} = (p_C^{\star}, \{p_E^{(i,\star)}\})$.
\State \textbf{// Phase 1: Coordinator optimization}
\For{$t = 1$ to $T_C$}
    \State $p_C^{(t)} \leftarrow \mathcal{O}\!\left(p_C^{(t-1)};\, \{p_E^{(i,0)}\}_{i=1}^{N},\, \Phi_C,\, \mathcal{D}_{\text{opt}}\right)$
\EndFor
\State $p_C^{\star} \leftarrow 
    \argmax_{t \in \{0, \ldots, T_C\}} J(p_C^{(t)},\, \{p_E^{(i,0)}\})$
\State \textbf{// Phase 2: Executor optimization}
\For{$i \in \mathcal{I}$}
    \For{$t = 1$ to $T_E$}
        \State $p_E^{(i,t)} \leftarrow \mathcal{O}\!\left(p_E^{(i,t-1)};\, p_C^{\star},\, \{p_E^{(j,0)}\}_{j \neq i},\, \Phi_E,\, \mathcal{D}_{\text{opt}}\right)$
    \EndFor
    \State $p_E^{(i,\star)} \leftarrow 
        \argmax_{t \in \{0, \ldots, T_E\}} J(p_C^{\star},\, p_E^{(i,t)})$
\EndFor
\For{$i \notin \mathcal{I}$} 
    \State $p_E^{(i,\star)} \leftarrow p_E^{(i,0)}$
\EndFor
\State \Return $(p_C^{\star},\, \{p_E^{(i,\star)}\}_{i=1}^{N})$
\end{algorithmic}
\end{algorithm}
\vspace{-0.4cm}

\section{Experiments}
\label{sec:experiments}
\subsection{Implementation Details}
\noindent\textbf{Benchmarks and evaluation.}
We evaluate CEPO on eleven benchmarks across four domains: document understanding (MMLongBench-Doc~\citep{ma2024mmlongbench}, LongDocURL~\citep{deng2025longdocurl}, and PaperTab, PaperText, and FetaTab~\citep{hui2024uda}), mathematical reasoning (AQuA~\citep{ling2017program}, MATH-500~\citep{lightman2024verify}, and AGIEval-MATH~\citep{zhong2024agieval}), code generation (HumanEval-ET~\citep{dong2025codescore} and MBPP~\citep{austin2021program}), and video understanding (LongTVQA+~\citep{liu2026longvideoagent}).

Since the document benchmarks lack dedicated training sets, we partition their test sets into document-level train/validation/test splits of 10/10/80 using a fixed seed. Baseline scores on our test splits are close to those on the original test sets (Appendix~\ref{sec:supp_dataset_statistics}). We evaluate answers with GPT-5-mini following the LLM-as-judge protocol of~\citep{han2025mdocagent} (Appendix~\ref{sec:evaluator_llm_prompt}). We follow MASPO~\citep{wang2026maspo} for mathematical reasoning and code generation settings. For video understanding, we follow LongVideoAgent~\citep{liu2026longvideoagent}, retaining its test set and reserving 50 questions from its training set for validation.

\noindent\textbf{Underlying prompt optimizers.}
For document experiments, we use PE2~\citep{ye2024prompt} with GPT-5.4 as the optimizer LLM and $T_C=T_E=3$. We also evaluate ProTeGi~\citep{pryzant2023automatic} and OPRO~\citep{yang2023large} on MDocAgent across all five benchmarks with the same LLM and iteration budgets, retaining each optimizer's default hyperparameters (Table~\ref{tab:optimization_algorithm}).
For mathematical reasoning and code generation, we follow the official MASPO~\citep{wang2026maspo} configurations. With MASPO, CEPO first optimizes the coordinator for up to $T_C=4$ rounds. Then, with the coordinator fixed, we perform two Predictor updates followed by two Reflector updates and repeat this sequence twice ($T_E=4$ per executor). GPT-5-mini generates and evaluates prompt candidates.
For video understanding, we use PE2 with GPT-5.4 and $T_C=T_E=3$, optimizing the coordinator first and then optimizing the grounding agent's executor prompt.

\noindent\textbf{Multi-agent systems.}
For document understanding, we use MDocAgent~\citep{han2025mdocagent} with Qwen-3.5 (4B)~\citep{team2026qwen3} and DocAgent~\citep{sun2025docagent} with GPT-5-mini.
For mathematical reasoning and code generation, we use MASPO's~\citep{wang2026maspo} Qwen3-8B-based MAS.
For video understanding, we use LongVideoAgent~\citep{liu2026longvideoagent} with Qwen2.5-7B as the master agent, and GPT-5-mini as grounding, and vision agents.

\noindent\textbf{Prompt initialization.}
For all evaluated multi-agent systems, executor prompts are initialized from the corresponding baseline's default prompts. 
An initial coordinator prompt is generated by GPT-5-mini from a description of the system and its agent roles. 
Examples in Tables~\ref{tab:seed_coordinator_prompt} and~\ref{tab:docagent_shared_seed_prompt}.

\subsection{Results}
\noindent\textbf{Document understanding.}
Table~\ref{tab:main} presents the results of applying PE2 + CEPO to MDocAgent and DocAgent across five document benchmarks. Compared to their unoptimized baselines, both systems improve on every benchmark, with average gains of 4.3 and 2.7 points, respectively. The largest gain occurs on MMLongBench-Doc with DocAgent, where the score rises from 59.5 to 67.4 (+7.9 points). These consistent gains across both agent systems and all five benchmarks demonstrate that the benefits extend across diverse document understanding tasks.

\begin{table*}[t]
\centering
\small
\setlength{\tabcolsep}{6pt}
\caption{\textbf{Results on Document understanding.} Results across five document understanding benchmarks. $^\dagger$: results on custom test split (Table~\ref{tab:split_performance_consistency}); others: results on original test set. ``--'': results not available. Avg.: average over benchmarks. Best results in \textbf{bold}, ours \colorbox{cyan!10}{highlighted}.}
\vspace{-0.3cm}

\begin{adjustbox}{width=\textwidth}
\begin{tabular}{lccccc|c}
\toprule
\textbf{Method} & \textbf{MMLongBench-Doc} & \textbf{LongDocURL} & \textbf{PaperTab} & \textbf{PaperText} & \textbf{FetaTab} & \textbf{Avg.} \\
\midrule
\rowcolor{lightgray}
\multicolumn{7}{l}{\textbf{\textit{Proprietary Models}}} \\
GPT-4.1$^\dagger$                & 49.0 & -- & -- & -- & -- & -- \\
Claude-4-Sonnet                & 53.4 & -- & -- & -- & -- & -- \\
Gemini-2.5-Pro                & 58.1 & -- & -- & -- & -- & -- \\
\rowcolor{lightgray}
\multicolumn{7}{l}{\textbf{\textit{Single-agent Systems}}} \\
M3DocRAG~\citep{cho2024m3docrag}                       & 29.6 & 50.3 & 15.2 & -- & 53.7 & -- \\
DocReact~\citep{wu2025doc}                       & 38.3 & -- & -- & -- & -- & -- \\
MoloRAG~\citep{wu2025molorag}                      & 37.1 & 53.6 & 15.7 & -- & 55.4 & -- \\
MoloRAG+~\citep{wu2025molorag}                      & 37.2 & 52.8 & 19.5 & -- & 60.0 & -- \\
MAB-DQA~\citep{xiang2026mab}                      & 39.9 & 56.4 & 26.9 & -- & 63.8 & -- \\
MLDocRAG~\citep{zhang2026mldocrag}                      & 47.9 & 50.8 & -- & -- & -- & -- \\
\rowcolor{lightgray}
\multicolumn{7}{l}{\textbf{\textit{Multi-agent Systems}}} \\
DMAP~\citep{fu2026dmap}                       & 43.2 & 60.7 & 39.0 & 56.7 & 72.5 & 54.4 \\
VisDoMRAG~\citep{suri2025visdom}                       & -- & -- & 44.1 & -- & 63.3 & -- \\
MDocAgent$^\dagger$~\citep{han2025mdocagent}                      & 54.1 & 74.9 & 46.3 & 72.1 & 78.8 & 65.2 \\
\rowcolor{cyan!10}
\ \ + \textbf{CEPO}$^\dagger$ (Ours)     
& 61.0$_{\textcolor{ForestGreen}{(+6.9)}}$ 
& \textbf{75.4}$_{\textcolor{ForestGreen}{(+0.5)}}$ 
& 53.7$_{\textcolor{ForestGreen}{(+7.4)}}$ 
& 76.4$_{\textcolor{ForestGreen}{(+4.3)}}$ 
& 80.8$_{\textcolor{ForestGreen}{(+2.0)}}$ 
& 69.5$_{\textcolor{ForestGreen}{(+4.3)}}$ \\
DocAgent$^\dagger$~\citep{sun2025docagent}                       
& 59.5 & 72.9 & 60.0 & 75.6 & 82.0 & 70.0 \\
\rowcolor{cyan!10}
\ \ + \textbf{CEPO}$^\dagger$ (Ours)     
& \textbf{67.4}$_{\textcolor{ForestGreen}{(+7.9)}}$ 
& 73.2$_{\textcolor{ForestGreen}{(+0.3)}}$ 
& \textbf{62.2}$_{\textcolor{ForestGreen}{(+2.2)}}$ 
& \textbf{78.3}$_{\textcolor{ForestGreen}{(+2.7)}}$ 
& \textbf{82.6}$_{\textcolor{ForestGreen}{(+0.6)}}$ 
& \textbf{72.7}$_{\textcolor{ForestGreen}{(+2.7)}}$ \\
\bottomrule
\end{tabular}
\end{adjustbox}
\label{tab:main}
\end{table*}

\begin{table*}[t]
\centering
\small
\setlength{\tabcolsep}{5pt}
\renewcommand{\arraystretch}{1.08}
\vspace{-0.3cm}
\caption{\textbf{CEPO with a multi-agent prompt optimizer.}
Results with MASPO~\citep{wang2026maspo} on mathematical reasoning and code generation. Rows are cumulative: + CEPO applies CEPO on top of MASPO. Green subscripts denote gains over the baseline. Avg.: average over benchmarks.}
\vspace{-0.25cm}

\label{tab:maspo}
\begin{adjustbox}{max width=\textwidth}
\begin{tabular}{@{}lccccc|c@{}}
\toprule
\multirow{2}{*}{\textbf{Method}}
& \multicolumn{3}{c}{\textbf{Math}}
& \multicolumn{2}{c|}{\textbf{Code}}
& \multirow{2}{*}{\textbf{Avg.}} \\
\cmidrule(lr){2-4}\cmidrule(lr){5-6}
& \textbf{AQuA} & \textbf{MATH-500} & \textbf{AGIEval-MATH}
& \textbf{HumanEval-ET} & \textbf{MBPP} & \\
\midrule
Baseline & 83.47 & 75.10 & 59.38 & 68.90 & 57.26 & 68.82 \\
\quad + MASPO
& 85.56 & 77.80 & 61.98 & 73.78 & 65.11 & 72.85 \\
\rowcolor{cyan!10}
\qquad + \textbf{CEPO}
& \textbf{87.01}$_{\textcolor{ForestGreen}{(+3.54)}}$
& \textbf{79.60}$_{\textcolor{ForestGreen}{(+4.50)}}$
& \textbf{62.89}$_{\textcolor{ForestGreen}{(+3.51)}}$
& \textbf{74.39}$_{\textcolor{ForestGreen}{(+5.49)}}$
& \textbf{65.57}$_{\textcolor{ForestGreen}{(+8.31)}}$
& \textbf{73.89}$_{\textcolor{ForestGreen}{(+5.07)}}$ \\
\bottomrule
\end{tabular}
\end{adjustbox}
\vspace{-0.4cm}
\end{table*}

\noindent\textbf{Mathematical reasoning and code generation.}
Table~\ref{tab:maspo} evaluates CEPO with MASPO~\citep{wang2026maspo} on mathematical reasoning and code generation. MASPO + CEPO outperforms both the baseline and MASPO alone on all five benchmarks, achieving an average gain of 5.07 points over the baseline. These results show that explicit coordinator-executor decomposition and role-aligned optimization can complement a prompt optimizer that is already tailored to multi-agent systems.

\noindent\textbf{Long video understanding.}
Table~\ref{tab:video} shows that applying CEPO on top of PE2 further improves LongVideoAgent~\citep{liu2026longvideoagent} on LongTVQA+ to 64.9, yielding a gain of 1.5 points over vanilla PE2 and 4.6 points over baseline, demonstrating that CEPO is effective for long-video understanding.

\begin{table}[t]
\centering
\small

\renewcommand{\arraystretch}{1.05}
\begin{minipage}[t]{0.48\textwidth}
\centering
\captionsetup{width=\linewidth}
\setlength{\tabcolsep}{4pt}
\caption{\textbf{Component ablation results.}}
\vspace{0.2cm}
\label{tab:component_ablation}
\begin{adjustbox}{max width=\linewidth}
\begin{tabular}{l|c}
\toprule
Configuration & Accuracy \\
\midrule
Baseline & 54.1 \\
\midrule
w/ Vanilla PE2 & 57.9 \\
\midrule
w/ Coordinator Prompt (LLM-generated) & 55.6 \\
w/ Coordinator Prompt (optimized) & 58.6 \\
\midrule
\rowcolor{cyan!10}
w/ Role-aligned Optimization & \textbf{61.0} \\
\bottomrule
\end{tabular}
\end{adjustbox}
\end{minipage}%
\hfill %
\begin{minipage}[t]{0.48\textwidth}
\centering
\captionsetup{width=\linewidth}
\setlength{\tabcolsep}{5pt}
\vspace{-0.25cm}
\caption{\textbf{Accuracy by training budget.} $T_C, T_E$: optimization iterations. Opt: external optimizer API calls. LLM: MAS internal API calls.}
\label{tab:budget}
\begin{tabular}{l|cc|cc|c}
\toprule
\multirow{2}{*}{Method} & \multirow{2}{*}{$T_C$} & \multirow{2}{*}{$T_E$} & \multicolumn{2}{c|}{\# API Calls} & \multirow{2}{*}{Acc.} \\
& & & Opt. & LLM & \\
\midrule
\multirow{2}{*}{PE2} & 0 & 3 & 10 & 5{,}714 & 57.9 \\
& 0 & 6 & 22 & 11{,}945 & 57.4 \\
\midrule
\multirow{2}{*}{+ CEPO} & 1 & 1 & 2 & 3{,}120 & 60.1 \\
& \cellcolor{cyan!10}3 & \cellcolor{cyan!10}3 & \cellcolor{cyan!10}20 & \cellcolor{cyan!10}11{,}458 & \cellcolor{cyan!10}\textbf{61.0} \\
\bottomrule
\end{tabular}
\end{minipage}
\vspace{-0.3cm}

\end{table}
\noindent\textbf{Component analysis.}
Table~\ref{tab:component_ablation} examines the prompt components on MDocAgent with MMLongBench-Doc, averaging results over three random seeds. Adding an LLM-generated coordinator without prompt optimization raises the score from 54.1 to 55.6. Optimizing only the executor with vanilla PE2 reaches 57.9, while optimizing only the coordinator reaches 58.6. PE2 + CEPO, which optimizes both the coordinator and the selected executor using role-aligned guidance, achieves 61.0. These comparisons support optimizing both shared coordination and local execution. The meta-instruction swap experiment separately examines the importance of matching update guidance to each component's role.

\subsection{Analysis}

Unless otherwise stated, the following analyses use PE2 with MDocAgent on MMLongBench-Doc. MDocAgent supports controlled changes to agent configurations and optimization targets, while MMLongBench-Doc includes diverse document reasoning cases.

\noindent\textbf{CEPO improves optimization budget efficiency.}
Table~\ref{tab:budget} shows that PE2 + CEPO with $T_C=T_E=1$ reaches 60.1, exceeding vanilla PE2 with three executor iterations by 2.2 points while using fewer optimizer and MAS API calls. At comparable API-call budgets, PE2 + CEPO with $T_C=T_E=3$ reaches 61.0, whereas vanilla PE2 with six executor iterations reaches 57.4. Additional controls examine inference cost (Table~\ref{tab:inf_cost}) and prompt length (Table~\ref{tab:length_controlled}): CEPO incurs little inference overhead in this document setting, and extending vanilla PE2's prompts to a comparable length does not recover CEPO's performance. These results support the value of how prompts are structured and optimized beyond simply increasing the optimization budget or prompt length.

\begin{wrapfigure}{r}{0.42\textwidth}
    \vspace{-0.4cm}
    \centering
    \includegraphics[width=\linewidth]{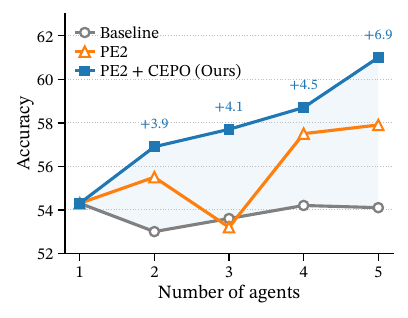}
    \vspace{-0.5cm}
    \caption{\textbf{Performance under varying numbers of agents.} Annotations indicate CEPO's gain over the baseline.}
    \vspace{-0.5cm}
    \label{fig:scaling}
    
\end{wrapfigure}
\noindent\textbf{CEPO improves performance scaling with MAS size.}
Figure~\ref{fig:scaling} compares the unoptimized system, vanilla PE2, and PE2 + CEPO on MDocAgent configurations containing one to five agents. The unoptimized system gains little as agents are added, while vanilla PE2 exhibits a non-monotonic trend and falls below the unoptimized result at three agents. PE2 + CEPO shows a monotonic increase in the reported scores, reaching 61.0 at five agents, and its advantage over the unoptimized system grows from 3.9 points at two agents to 6.9 at five agents. With two agents, it already exceeds the best unoptimized five-agent result. These observations support the benefit of explicit coordination optimization as the number of collaborating agents increases in this system. Per-configuration results and the aggregation procedure are provided in Appendix~\ref{sec:config_n_agent}.

\begin{table}[t]
\centering
\small
\begin{minipage}[t]{0.48\linewidth}
\centering
\setlength{\tabcolsep}{8pt}
\caption{\textbf{Optimization guidance swap experiment.} Role-aligned guidance ($p_C$ = coordination, $p_E$ = execution) perform best, and reversing them performs worst.}
\vspace{-0.3cm}
\begin{tabular}{c|cc}
\toprule
\diagbox[width=2.2cm]{$p_C$}{$p_E$} & \textsc{Coord} & \textsc{Exec} \\
\midrule
\textsc{Coord} & 58.1 & \cellcolor{cyan!10}\textbf{61.0} \\
\textsc{Exec} & 57.7 & 59.1 \\
\bottomrule
\end{tabular}
\label{tab:swap}
\end{minipage}
\hfill
\begin{minipage}[t]{0.48\linewidth}
\centering
\setlength{\tabcolsep}{6pt}
\renewcommand{\arraystretch}{1.08}
\caption{\textbf{Results on long-video understanding.}
Rows are cumulative.
The green subscript denotes the gain over the baseline.}
\label{tab:video}
\begin{tabular}{@{}lc@{}}
\toprule
\textbf{Method} & \textbf{LongTVQA+} \\
\midrule
LongVideoAgent~\citep{liu2026longvideoagent} & 60.3 \\
\quad + PE2 & 63.4 \\
\rowcolor{cyan!10}
\qquad + \textbf{CEPO} & \textbf{64.9}$_{\textcolor{ForestGreen}{(+4.6)}}$ \\
\bottomrule
\end{tabular}
\end{minipage}
\vspace{-0.3cm}
\end{table}

\begin{table}[t]
\centering
\small
\setlength{\tabcolsep}{4pt}
\renewcommand{\arraystretch}{1.0}
\caption{\textbf{CEPO with different prompt optimizers across document benchmarks.}
Results with PE2, ProTeGi, and OPRO on MDocAgent across five document benchmarks.
Within each optimizer block, rows are cumulative: + CEPO applies CEPO on top of the corresponding optimizer.}
\vspace{-0.3cm}
\label{tab:optimization_algorithm}
\begin{adjustbox}{max width=\linewidth}
\begin{tabular}{@{}lccccc@{}}
\toprule
\textbf{Method} & \textbf{MMLongBench-Doc} & \textbf{LongDocURL} & \textbf{PaperTab} & \textbf{PaperText} & \textbf{FetaTab} \\
\midrule
MDocAgent & 54.1 & 74.9 & 46.3 & 72.1 & 78.8 \\
\midrule
\quad + PE2 & 57.9 & 74.4 & 48.1 & 71.4 & 77.8 \\
\rowcolor{cyan!10}
\qquad + \textbf{CEPO} & \textbf{61.0} & \textbf{75.4} & \textbf{53.7} & \textbf{76.4} & \textbf{80.8} \\
\midrule
\quad + ProTeGi & 55.5 & 73.5 & 46.6 & 72.7 & 78.4 \\
\rowcolor{cyan!10}
\qquad + \textbf{CEPO} & \textbf{57.9} & \textbf{75.4} & \textbf{53.1} & \textbf{76.7} & \textbf{79.2} \\
\midrule
\quad + OPRO & 57.3 & 73.2 & 46.3 & 70.9 & 78.9 \\
\rowcolor{cyan!10}
\qquad + \textbf{CEPO} & \textbf{60.4} & \textbf{75.2} & \textbf{46.9} & \textbf{73.0} & \textbf{80.0} \\
\bottomrule
\end{tabular}
\end{adjustbox}
\vspace{-0.4cm}

\end{table}

\noindent\textbf{Role-aligned guidance improves optimization.}
With the coordinator-executor decomposition fixed, we compare four assignments of optimization guidance (Table~\ref{tab:swap}). Applying coordination guidance $\Phi_C$ to the coordinator and execution guidance $\Phi_E$ to the executor achieves the best score of 61.0, whereas reversing the assignment yields the worst score of 57.7. Using the same guidance for both layers gives intermediate results: 58.1 (coordination guidance) and 59.1 (execution guidance). These results show the importance of using distinct optimization guidance aligned with each layer's role.

\noindent\textbf{CEPO complements different prompt optimizers.}
Table~\ref{tab:optimization_algorithm} compares PE2~\citep{ye2024prompt}, ProTeGi~\citep{pryzant2023automatic}, and OPRO~\citep{yang2023large} with and without CEPO on MDocAgent across all five document benchmarks. CEPO improves performance in every optimizer-benchmark combination, with gains of 0.6 to 6.5 points over the corresponding vanilla optimizer. Table~\ref{tab:maspo} further shows that CEPO improves MASPO~\citep{wang2026maspo}, a multi-agent prompt optimizer, on all five mathematical reasoning and code generation benchmarks. Together, these results demonstrate that CEPO complements both general-purpose and multi-agent prompt optimizers.

\begin{figure*}[t!]
    \centering
    \includegraphics[width=\textwidth]{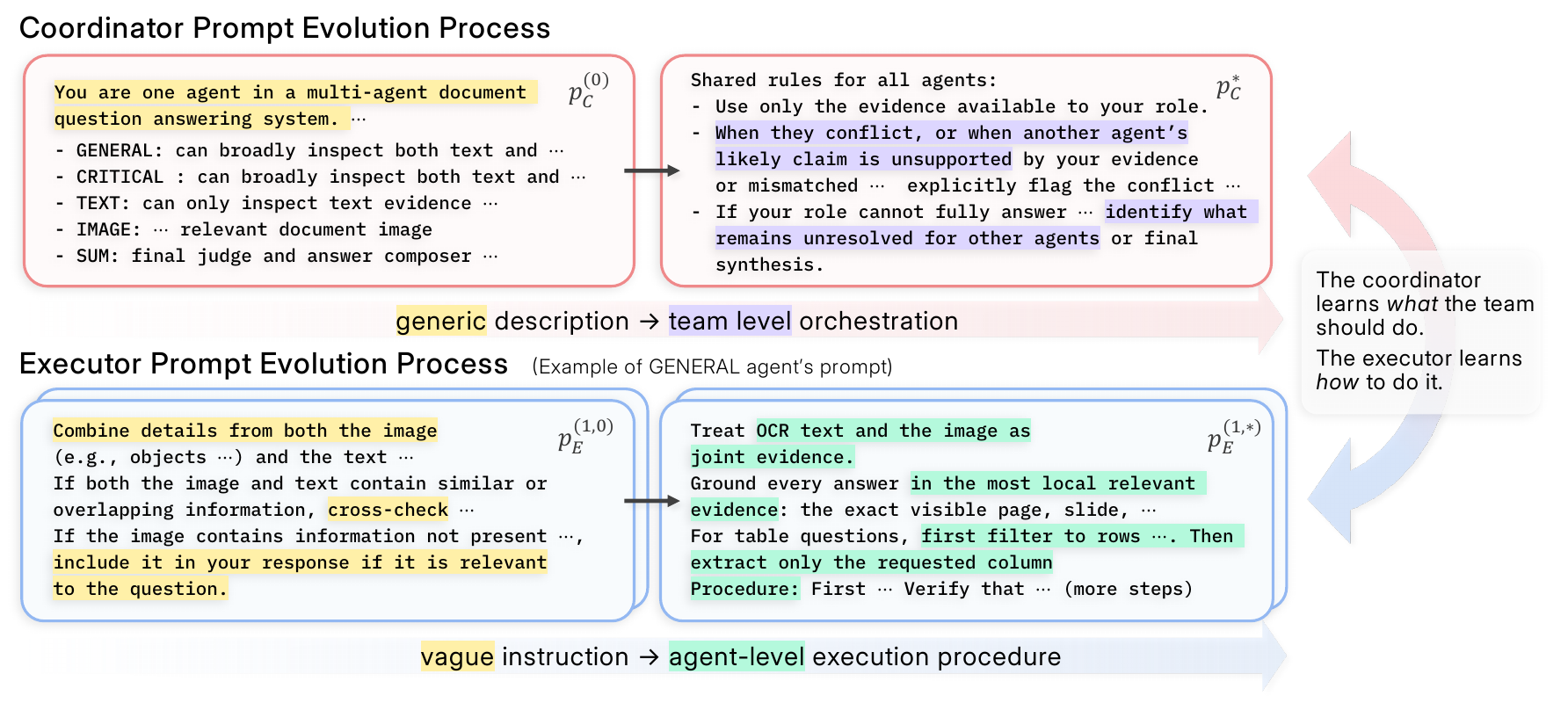}
    \vspace{-0.6cm}
    \caption{\textbf{Qualitative comparison of optimized prompts.} 
Through optimization, the coordinator prompt evolves from a generic description into team-level orchestration rules. The executor prompt of \textsc{GENERAL} agent evolves from vague instructions into a concrete agent-level execution procedure.}
\vspace{-0.5cm}
 
\label{fig:qualitative}
\label{fig:qual}
\end{figure*}

\noindent\textbf{Coordinator and executor prompts develop complementary roles.}
Figure~\ref{fig:qualitative} shows how role-aligned guidance transforms a generic system description into detailed team-level coordination and vague executor instructions into concrete agent-level execution procedures. The coordinator adds shared rules for flagging unsupported claims, identifying unresolved issues, and supporting final synthesis. The executor specifies how to combine OCR and image evidence, ground answers, and extract table information. These examples illustrate the distinction between shared coordination and agent-specific execution under role-aligned guidance.

\section{Conclusion}
\label{sec:conclusion}
We presented CEPO, a general framework that addresses a shared limitation in MAS prompt optimization: coupled coordination and execution. By combining coordinator-executor prompt decomposition with role-aligned optimization, CEPO enables existing optimizers to refine both components considering their respective roles. Applying CEPO to four different prompt optimizers yields consistent gains in extensive experiments spanning eleven benchmarks across four domains. Controlled studies further support the proposed design and show improved optimization budget efficiency and performance scaling with MAS size.

\subsection*{AI use statement}
LLMs are used as components of our experimental framework, including prompt optimization and LLM-as-judge evaluation, as described in the paper. We also used generative AI tools to assist with coding during experiments, to identify and retrieve relevant literature, and to improve the clarity and language of the manuscript. The authors take responsibility for the final manuscript, code, analyses, and reported results.

\subsection*{Ethics statement}
CEPO can reduce the manual effort required to design prompts for multi-agent systems, making these systems more accessible. However, easier prompt optimization may encourage deployment without sufficient validation. Optimized prompts can still produce unintended model behavior, particularly in high-stakes or out-of-distribution settings. We therefore recommend task-specific validation and appropriate human oversight before deployment.

\subsection*{Reproducibility statement}
Section~\ref{sec:method} presents CEPO and its optimization procedure. Section~4.1 and Appendix~\ref{sec:supp_implementation_details} describe the experimental setup, dataset splits, model configurations, and computing resources. Appendix~\ref{sec:prompts} provides evaluator prompts, optimization meta-instructions, and examples of initial and optimized prompts. We plan to release the full codebase upon acceptance.

{
    \small
    \bibliographystyle{iclr2027_conference}
\bibliography{main}

\begin{thebibliography}{45}
\providecommand{\natexlab}[1]{#1}
\providecommand{\url}[1]{\texttt{#1}}
\expandafter\ifx\csname urlstyle\endcsname\relax
  \providecommand{\doi}[1]{doi: #1}\else
  \providecommand{\doi}{doi: \begingroup \urlstyle{rm}\Url}\fi

\bibitem[Achiam et~al.(2023)Achiam, Adler, Agarwal, Ahmad, Akkaya, Aleman,
  Almeida, Altenschmidt, Altman, Anadkat, et~al.]{achiam2023gpt}
Josh Achiam, Steven Adler, Sandhini Agarwal, Lama Ahmad, Ilge Akkaya,
  Florencia~Leoni Aleman, Diogo Almeida, Janko Altenschmidt, Sam Altman,
  Shyamal Anadkat, et~al.
\newblock {GPT-4} technical report.
\newblock \emph{arXiv preprint arXiv:2303.08774}, 2023.

\bibitem[Austin et~al.(2021)Austin, Odena, Nye, Bosma, Michalewski, Dohan,
  Jiang, Cai, Terry, Le, and Sutton]{austin2021program}
Jacob Austin, Augustus Odena, Maxwell Nye, Maarten Bosma, Henryk Michalewski,
  David Dohan, Ellen Jiang, Carrie Cai, Michael Terry, Quoc Le, and Charles
  Sutton.
\newblock Program synthesis with large language models.
\newblock \emph{arXiv preprint arXiv:2108.07732}, 2021.

\bibitem[Cho et~al.(2024)Cho, Mahata, Irsoy, He, and Bansal]{cho2024m3docrag}
Jaemin Cho, Debanjan Mahata, Ozan Irsoy, Yujie He, and Mohit Bansal.
\newblock {M3DocRAG}: Multi-modal retrieval is what you need for multi-page
  multi-document understanding.
\newblock \emph{arXiv preprint arXiv:2411.04952}, 2024.

\bibitem[Deng et~al.(2025)Deng, Yuan, Bu, Wang, Li, Xu, Li, Gao, Song, Zheng,
  et~al.]{deng2025longdocurl}
Chao Deng, Jiale Yuan, Pi~Bu, Peijie Wang, Zhong-Zhi Li, Jian Xu, Xiao-Hui Li,
  Yuan Gao, Jun Song, Bo~Zheng, et~al.
\newblock Longdocurl: a comprehensive multimodal long document benchmark
  integrating understanding, reasoning, and locating.
\newblock In \emph{ACL}, 2025.

\bibitem[Dong et~al.(2025)Dong, Ding, Jiang, Li, Li, and
  Jin]{dong2025codescore}
Yihong Dong, Jiazheng Ding, Xue Jiang, Ge~Li, Zhuo Li, and Zhi Jin.
\newblock {CodeScore}: Evaluating code generation by learning code execution.
\newblock \emph{ACM Transactions on Software Engineering and Methodology},
  34\penalty0 (3), 2025.
\newblock \doi{10.1145/3695991}.

\bibitem[Du et~al.(2024)Du, Li, Torralba, Tenenbaum, and
  Mordatch]{du2024improving}
Yilun Du, Shuang Li, Antonio Torralba, Joshua~B. Tenenbaum, and Igor Mordatch.
\newblock Improving factuality and reasoning in language models through
  multiagent debate.
\newblock In \emph{ICML}, 2024.

\bibitem[Fu et~al.(2026)Fu, Zhang, Xiang, Du, and Tang]{fu2026dmap}
ShunLiang Fu, Yanxin Zhang, Yixin Xiang, Xiaoyu Du, and Jinhui Tang.
\newblock Dmap: Human-aligned structural document map for multimodal document
  understanding.
\newblock In \emph{Proceedings of the ACM Web Conference 2026}, 2026.

\bibitem[Guo et~al.(2024)Guo, Wang, Guo, Li, Song, Tan, Liu, Bian, and
  Yang]{guo2023evoprompt}
Qingyan Guo, Rui Wang, Junliang Guo, Bei Li, Kaitao Song, Xu~Tan, Guoqing Liu,
  Jiang Bian, and Yujiu Yang.
\newblock Connecting large language models with evolutionary algorithms yields
  powerful prompt optimizers.
\newblock In \emph{ICLR}, 2024.

\bibitem[Han et~al.(2025{\natexlab{a}})Han, Xia, Zhang, Sun, Li, Zhu, and
  Yao]{han2025mdocagent}
Siwei Han, Peng Xia, Ruiyi Zhang, Tong Sun, Yun Li, Hongtu Zhu, and Huaxiu Yao.
\newblock Mdocagent: A multi-modal multi-agent framework for document
  understanding.
\newblock \emph{arXiv preprint arXiv:2503.13964}, 2025{\natexlab{a}}.

\bibitem[Han et~al.(2025{\natexlab{b}})Han, Han, Liu, Zhou, Liu, Zhang, Yang,
  Wang, Shi, Zhang, He, and Shi]{han2025mapgdmultiagentpromptgradient}
Yichen Han, Yuhang Han, Bojun Liu, Zhengpeng Zhou, Guanyu Liu, Zeng Zhang, Yang
  Yang, Wenli Wang, Isaac~N Shi, Yunyan Zhang, Lewei He, and Tianyu Shi.
\newblock Mapgd: Multi-agent prompt gradient descent for collaborative prompt
  optimization, 2025{\natexlab{b}}.
\newblock URL \url{https://arxiv.org/abs/2509.11361}.

\bibitem[Hong et~al.(2026)Hong, Zhang, Sun, Shang, Kong, Wang, Shu, and
  Dai]{hong2026maspob}
Zhi Hong, Qian Zhang, Jiahang Sun, Zhiwei Shang, Mingze Kong, Xiangyi Wang, Yao
  Shu, and Zhongxiang Dai.
\newblock Maspob: Bandit-based prompt optimization for multi-agent systems with
  graph neural networks.
\newblock \emph{arXiv preprint arXiv:2603.02630}, 2026.

\bibitem[Huang et~al.(2023{\natexlab{a}})Huang, Zhang, Luck, Bu, Qing, and
  Cui]{huang2023agentcoder}
Dong Huang, Jie~M. Zhang, Michael Luck, Qingwen Bu, Yuhao Qing, and Heming Cui.
\newblock {AgentCoder}: Multi-agent-based code generation with iterative
  testing and optimisation.
\newblock \emph{arXiv preprint arXiv:2312.13010}, 2023{\natexlab{a}}.

\bibitem[Huang et~al.(2023{\natexlab{b}})Huang, Dong, Wang, Hao, Singhal, Ma,
  Lv, Cui, Mohammed, Patra, et~al.]{huang2023language}
Shaohan Huang, Li~Dong, Wenhui Wang, Yaru Hao, Saksham Singhal, Shuming Ma,
  Tengchao Lv, Lei Cui, Owais~Khan Mohammed, Barun Patra, et~al.
\newblock Language is not all you need: Aligning perception with language
  models.
\newblock In \emph{NeurIPS}, 2023{\natexlab{b}}.

\bibitem[Hui et~al.(2024)Hui, Lu, and Zhang]{hui2024uda}
Yulong Hui, Yao Lu, and Huanchen Zhang.
\newblock {UDA}: A benchmark suite for retrieval augmented generation in
  real-world document analysis.
\newblock In \emph{NeurIPS}, 2024.

\bibitem[Islam et~al.(2024)Islam, Ali, and Parvez]{islam2024mapcoder}
Md.~Ashraful Islam, Mohammed~Eunus Ali, and Md~Rizwan Parvez.
\newblock {MapCoder}: Multi-agent code generation for competitive problem
  solving.
\newblock In \emph{ACL}, 2024.

\bibitem[Khattab et~al.(2024)Khattab, Singhvi, Maheshwari, Zhang, Santhanam,
  Vardhamanan, Haq, Sharma, Joshi, Moazam, Miller, Zaharia, and
  Potts]{khattab2024dspy}
Omar Khattab, Arnav Singhvi, Paridhi Maheshwari, Zhiyuan Zhang, Keshav
  Santhanam, Sri Vardhamanan, Saiful Haq, Ashutosh Sharma, Thomas~T. Joshi,
  Hanna Moazam, Heather Miller, Matei Zaharia, and Christopher Potts.
\newblock Dspy: Compiling declarative language model calls into self-improving
  pipelines.
\newblock In \emph{ICLR}, 2024.

\bibitem[Kugo et~al.(2025)Kugo, Li, Li, Gupta, Khatua, Jain, Patel, Kyuragi,
  Ishii, Tanabiki, Kozuka, and Adeli]{kugo2025videomultiagents}
Noriyuki Kugo, Xiang Li, Zixin Li, Ashish Gupta, Arpandeep Khatua, Nidhish
  Jain, Chaitanya Patel, Yuta Kyuragi, Yasunori Ishii, Masamoto Tanabiki,
  Kazuki Kozuka, and Ehsan Adeli.
\newblock {VideoMultiAgents}: A multi-agent framework for video question
  answering.
\newblock \emph{arXiv preprint arXiv:2504.20091}, 2025.

\bibitem[Li et~al.(2023)Li, Hammoud, Itani, Khizbullin, and
  Ghanem]{li2023camel}
Guohao Li, Hasan Hammoud, Hani Itani, Dmitrii Khizbullin, and Bernard Ghanem.
\newblock Camel: Communicative agents for" mind" exploration of large language
  model society.
\newblock In \emph{NeurIPS}, 2023.

\bibitem[Lightman et~al.(2024)Lightman, Kosaraju, Burda, Edwards, Baker, Lee,
  Leike, Schulman, Sutskever, and Cobbe]{lightman2024verify}
Hunter Lightman, Vineet Kosaraju, Yuri Burda, Harrison Edwards, Bowen Baker,
  Teddy Lee, Jan Leike, John Schulman, Ilya Sutskever, and Karl Cobbe.
\newblock Let's verify step by step.
\newblock In \emph{ICLR}, 2024.

\bibitem[Ling et~al.(2017)Ling, Yogatama, Dyer, and Blunsom]{ling2017program}
Wang Ling, Dani Yogatama, Chris Dyer, and Phil Blunsom.
\newblock Program induction by rationale generation: Learning to solve and
  explain algebraic word problems.
\newblock In \emph{ACL}, 2017.

\bibitem[Liu et~al.(2026)Liu, Liu, Tang, Ma, Pi, Zhang, and
  Chen]{liu2026longvideoagent}
Runtao Liu, Ziyi Liu, Jiaqi Tang, Yue Ma, Renjie Pi, Jipeng Zhang, and Qifeng
  Chen.
\newblock {LongVideoAgent}: Multi-agent reasoning with long videos.
\newblock In \emph{ACL}, 2026.

\bibitem[Ma et~al.(2024)Ma, Zang, Chen, Chen, Jiao, Li, Lu, Liu, Ma, Dong,
  et~al.]{ma2024mmlongbench}
Yubo Ma, Yuhang Zang, Liangyu Chen, Meiqi Chen, Yizhu Jiao, Xinze Li, Xinyuan
  Lu, Ziyu Liu, Yan Ma, Xiaoyi Dong, et~al.
\newblock {MMLongBench-Doc}: Benchmarking long-context document understanding
  with visualizations.
\newblock In \emph{NeurIPS}, 2024.

\bibitem[Opsahl-Ong et~al.(2024)Opsahl-Ong, Ryan, Purtell, Broman, Potts,
  Zaharia, and Khattab]{opsahl2024optimizing}
Krista Opsahl-Ong, Michael~J Ryan, Josh Purtell, David Broman, Christopher
  Potts, Matei Zaharia, and Omar Khattab.
\newblock Optimizing instructions and demonstrations for multi-stage language
  model programs.
\newblock In \emph{EMNLP}, 2024.

\bibitem[Pryzant et~al.(2023)Pryzant, Iter, Li, Lee, Zhu, and
  Zeng]{pryzant2023automatic}
Reid Pryzant, Dan Iter, Jerry Li, Yin Lee, Chenguang Zhu, and Michael Zeng.
\newblock Automatic prompt optimization with ``gradient descent'' and beam
  search.
\newblock In \emph{EMNLP}, 2023.

\bibitem[{Qwen Team}(2026)]{team2026qwen3}
{Qwen Team}.
\newblock Qwen3.5: Towards native multimodal agents.
\newblock \url{https://qwen.ai/blog?id=qwen3.5}, February 2026.

\bibitem[Schick et~al.(2023)Schick, Dwivedi-Yu, Dess{\`\i}, Raileanu, Lomeli,
  Hambro, Zettlemoyer, Cancedda, and Scialom]{schick2023toolformer}
Timo Schick, Jane Dwivedi-Yu, Roberto Dess{\`\i}, Roberta Raileanu, Maria
  Lomeli, Eric Hambro, Luke Zettlemoyer, Nicola Cancedda, and Thomas Scialom.
\newblock Toolformer: Language models can teach themselves to use tools.
\newblock In \emph{NeurIPS}, 2023.

\bibitem[Sun et~al.(2025)Sun, He, Jia, He, and You]{sun2025docagent}
Li~Sun, Liu He, Shuyue Jia, Yangfan He, and Chenyu You.
\newblock {D}oc{A}gent: An agentic framework for multi-modal long-context
  document understanding.
\newblock In \emph{EMNLP}, 2025.

\bibitem[Suri et~al.(2025)Suri, Mathur, Dernoncourt, Goswami, Rossi, and
  Manocha]{suri2025visdom}
Manan Suri, Puneet Mathur, Franck Dernoncourt, Kanika Goswami, Ryan~A Rossi,
  and Dinesh Manocha.
\newblock Visdom: Multi-document qa with visually rich elements using
  multimodal retrieval-augmented generation.
\newblock In \emph{NAACL}, 2025.

\bibitem[Wang et~al.(2026{\natexlab{a}})Wang, Chen, Qin, Wei, Qian, and
  Bai]{wang2026think}
Zheng Wang, Haoran Chen, Haoxuan Qin, Zhipeng Wei, Tianwen Qian, and Cong Bai.
\newblock Think, then verify: A hypothesis-verification multi-agent framework
  for long video understanding.
\newblock In \emph{CVPR}, 2026{\natexlab{a}}.

\bibitem[Wang et~al.(2026{\natexlab{b}})Wang, Liu, Wang, Shan, Wang, Song, and
  Zhang]{wang2026maspo}
Zhexuan Wang, Xuebo Liu, Li~Wang, Zifei Shan, Yutong Wang, Zhenxi Song, and Min
  Zhang.
\newblock {MASPO}: Joint prompt optimization for {LLM}-based multi-agent
  systems, 2026{\natexlab{b}}.
\newblock URL \url{https://arxiv.org/abs/2605.06623}.

\bibitem[Wu et~al.(2025{\natexlab{a}})Wu, Xia, Yu, Chen, Harsha, Maharaj,
  Zhang, Bursztyn, Kim, Rossi, et~al.]{wu2025doc}
Junda Wu, Yu~Xia, Tong Yu, Xiang Chen, Sai~Sree Harsha, Akash~V Maharaj, Ruiyi
  Zhang, Victor Bursztyn, Sungchul Kim, Ryan~A Rossi, et~al.
\newblock Doc-react: Multi-page heterogeneous document question-answering.
\newblock In \emph{ACL (short)}, 2025{\natexlab{a}}.

\bibitem[Wu et~al.(2024)Wu, Bansal, Zhang, Wu, Li, Zhu, Jiang, Zhang, Zhang,
  Liu, et~al.]{wu2024autogen}
Qingyun Wu, Gagan Bansal, Jieyu Zhang, Yiran Wu, Beibin Li, Erkang Zhu,
  Li~Jiang, Xiaoyun Zhang, Shaokun Zhang, Jiale Liu, et~al.
\newblock Autogen: Enabling next-gen llm applications via multi-agent
  conversations.
\newblock In \emph{COLM}, 2024.

\bibitem[Wu et~al.(2025{\natexlab{b}})Wu, Tan, Hou, Zhang, and
  Cheng]{wu2025molorag}
Xixi Wu, Yanchao Tan, Nan Hou, Ruiyang Zhang, and Hong Cheng.
\newblock {MoloRAG}: Bootstrapping document understanding via multi-modal
  logic-aware retrieval.
\newblock In \emph{EMNLP}, 2025{\natexlab{b}}.

\bibitem[Xiang et~al.(2026)Xiang, Ma, Du, Chen, Zhang, and Tang]{xiang2026mab}
Yixin Xiang, Yunshan Ma, Xiaoyu Du, Yibing Chen, Yanxin Zhang, and Jinhui Tang.
\newblock Mab-dqa: Addressing query aspect importance in document question
  answering with multi-armed bandits.
\newblock In \emph{ACL}, 2026.

\bibitem[Yang et~al.(2024)Yang, Wang, Lu, Liu, Le, Zhou, and
  Chen]{yang2023large}
Chengrun Yang, Xuezhi Wang, Yifeng Lu, Hanxiao Liu, Quoc~V Le, Denny Zhou, and
  Xinyun Chen.
\newblock Large language models as optimizers.
\newblock In \emph{ICLR}, 2024.

\bibitem[Ye et~al.(2024)Ye, Ahmed, Pryzant, and Khani]{ye2024prompt}
Qinyuan Ye, Mohamed Ahmed, Reid Pryzant, and Fereshte Khani.
\newblock Prompt engineering a prompt engineer.
\newblock In \emph{ACL (Findings)}, 2024.

\bibitem[Yuan \& Xie(2025)Yuan and Xie]{yuan2025reinforce}
Yurun Yuan and Tengyang Xie.
\newblock Reinforce {LLM} reasoning through multi-agent reflection.
\newblock In \emph{ICML}, 2025.

\bibitem[Yuksekgonul et~al.(2025)Yuksekgonul, Bianchi, Boen, Liu, Lu, Huang,
  Guestrin, and Zou]{yuksekgonul2025optimizing}
Mert Yuksekgonul, Federico Bianchi, Joseph Boen, Sheng Liu, Pan Lu, Zhi Huang,
  Carlos Guestrin, and James Zou.
\newblock Optimizing generative ai by backpropagating language model feedback.
\newblock \emph{Nature}, 639, 2025.

\bibitem[Zhang \& Wu(2026)Zhang and Wu]{zhang2026mldocrag}
Yongyue Zhang and Yaxiong Wu.
\newblock Mldocrag: Multimodal long-context document retrieval augmented
  generation.
\newblock \emph{arXiv preprint arXiv:2602.10271}, 2026.

\bibitem[Zhang et~al.(2026)Zhang, Ge, Li, Zhu, Zhang, and Ye]{zhang2026mapro}
Zheyuan Zhang, Lin Ge, Hongjiang Li, Weicheng Zhu, Chuxu Zhang, and Yanfang Ye.
\newblock Mapro: Recasting multi-agent prompt optimization as maximum a
  posteriori inference.
\newblock In \emph{EACL (Findings)}, 2026.

\bibitem[Zhong et~al.(2024)Zhong, Cui, Guo, Liang, Lu, Wang, Saied, Chen, and
  Duan]{zhong2024agieval}
Wanjun Zhong, Ruixiang Cui, Yiduo Guo, Yaobo Liang, Shuai Lu, Yanlin Wang, Amin
  Saied, Weizhu Chen, and Nan Duan.
\newblock {AGIEval}: A human-centric benchmark for evaluating foundation
  models.
\newblock In \emph{NAACL (Findings)}, 2024.

\bibitem[Zhou et~al.(2026)Zhou, Wan, Sun, Palangi, Iqbal, Vuli{\'c}, Korhonen,
  and Ar{\i}k]{zhou2025multi}
Han Zhou, Xingchen Wan, Ruoxi Sun, Hamid Palangi, Shariq Iqbal, Ivan Vuli{\'c},
  Anna Korhonen, and Sercan~{\"O} Ar{\i}k.
\newblock Multi-agent design: Optimizing agents with better prompts and
  topologies.
\newblock In \emph{ICLR}, 2026.

\bibitem[Zhou et~al.(2023)Zhou, Muresanu, Han, Paster, Pitis, Chan, and
  Ba]{zhou2022large}
Yongchao Zhou, Andrei~Ioan Muresanu, Ziwen Han, Keiran Paster, Silviu Pitis,
  Harris Chan, and Jimmy Ba.
\newblock Large language models are human-level prompt engineers.
\newblock In \emph{ICLR}, 2023.

\bibitem[Zhu et~al.(2025)Zhu, Meng, Chen, Li, Pfister, and
  Yoon]{zhu2025doclens}
Dawei Zhu, Rui Meng, Jiefeng Chen, Sujian Li, Tomas Pfister, and Jinsung Yoon.
\newblock Doclens : A tool-augmented multi-agent framework for long visual
  document understanding.
\newblock \emph{arXiv preprint arXiv:2511.11552}, 2025.

\bibitem[Zhuge et~al.(2024)Zhuge, Wang, Kirsch, Faccio, Khizbullin, and
  Schmidhuber]{zhuge2024gptswarm}
Mingchen Zhuge, Wenyi Wang, Louis Kirsch, Francesco Faccio, Dmitrii Khizbullin,
  and J{\"u}rgen Schmidhuber.
\newblock {GPTSwarm}: Language agents as optimizable graphs.
\newblock In \emph{ICML}, 2024.

\end{thebibliography}
}

\clearpage
\appendix

\setcounter{table}{0}
\renewcommand{\thetable}{S\arabic{table}}
\setcounter{figure}{0}
\renewcommand{\thefigure}{S\arabic{figure}}

\section{Experimental Details}
\label{sec:supp_implementation_details}
\subsection{Dataset statistics}
\label{sec:supp_dataset_statistics}
We split each dataset at the document level with a 10/10/80 train/validation/test ratio: documents are randomly shuffled and assigned to splits, and all questions associated with a given document are kept together. 
Since many documents contain multiple questions (Table~\ref{tab:doc_level_split_stats}), a question-level split would leak shared document context across splits. 
As shown in Table~\ref{tab:split_performance_consistency}, performance on our custom test split closely matches that on the original test split, confirming that our split does not introduce an easier evaluation. 
For mathematical reasoning and code generation, we use the same datasets and splits as MASPO~\citep{wang2026maspo}. For LongTVQA+, we use the same test set as LongVideoAgent~\citep{liu2026longvideoagent}. We hold out 50 questions from the original training set for validation and use the remainder as our training set.
\subsection{Computing resource}
For MDocAgent, the agent backbone is Qwen-3.5 (4B)~\citep{team2026qwen3}, served locally on 8$\times$H200 GPUs. For DocAgent, the agent backbone is GPT-5-mini, accessed through the OpenAI API. 
For the video experiments, we use Qwen2.5-7B as the locally served master agent, while the grounding and vision agents are powered by GPT-5-mini through the OpenAI API. We use GPT-5.4 as the optimizer LLM.

\subsection{Further implementation details}
We instantiate CEPO on two competitive multi-agent systems for document understanding. 
MDocAgent~\citep{han2025mdocagent} is a five-agent multi-agent system (\textsc{GENERAL}, \textsc{CRITIC}, \textsc{TEXT}, \textsc{IMAGE}, \textsc{SUM}) that splits processing across text and image modalities: the \textsc{GENERAL} and \textsc{CRITIC} agents produce initial answer and identifies crucial clues, the \textsc{TEXT} and \textsc{IMAGE} agents process modality-specific evidence, and the \textsc{SUM} agent aggregates all intermediate outputs into the final answer. 
DocAgent~\citep{sun2025docagent} is a three-agent actor-reviewer-reflector.
The actor drafts an answer, the reviewer critiques it, and the reflector revises the answer based on the critique. 
Agents in MDocAgent and DocAgent are powered by Qwen-3.5 (4B)~\citep{team2026qwen3} and GPT-5-mini, respectively.
Executor prompts are initialized from the default prompts published with each system, and the coordinator prompt for CEPO is initialized by querying GPT-5-mini with a brief description of the agent composition (Appendix~\ref{sec:llm_gen_coord_prompt}).

\subsection{Configuration by number of agents}
Table~\ref{tab:agent_scaling} reports the per-configuration breakdown underlying the agent-count analysis in the main text. From the full set of 5 agent roles (\textsc{GENERAL}, \textsc{CRITIC}, \textsc{TEXT}, \textsc{IMAGE}, \textsc{SUM}), we construct subsets of size 1 to 5 and retain only those configurations that produce a valid final answer. 
When multiple valid configurations exist for a given agent count, we report each configuration's accuracy individually and provide the arithmetic mean across them as the headline result; the bold row in each group indicates the value used for cross-group comparison. 
We compare three settings: baseline (unoptimized prompts), PE2, and PE2 + CEPO.
Across all agent counts and individual configurations, CEPO consistently outperforms both baseline and vanilla PE2, with the margin over baseline growing from 3.9 at 2 agents to 6.9 at 5 agents. 
Vanilla PE2 is less stable, even falling slightly below baseline at 3 agents due to a sharp drop on \textsc{GENERAL} + \textsc{TEXT} + \textsc{SUM}.

\label{sec:config_n_agent}

\begin{table}[t]
\centering
\small
\setlength{\tabcolsep}{6pt}
\caption{
\textbf{Test set accuracy on full set and split set.}
$\Delta$: Split test set accuracy - full test set accuracy.
}
\label{tab:split_performance_consistency}
\begin{tabular}{lccc}
\toprule
Dataset & Full test set accuracy & Split test set accuracy & $\Delta$ \\
\midrule
MMLongBench-Doc  & 54.2 & 54.1 & -0.1 \\
LongDocURL   & 73.1 & 74.9 & 1.8 \\
PaperTab     & 47.6 & 46.3 & -1.3 \\
PaperText    & 71.1 & 72.1 & 1.0 \\
FetaTab      & 77.2 & 78.8 & 1.6 \\
\bottomrule
\end{tabular}
\end{table}
\begin{table*}[t]
\centering
\small
\setlength{\tabcolsep}{5.5pt}
\caption{
\textbf{Data split statistics.}
Each split is constructed at the PDF/document level. \# Q denotes the number of questions.
}
\label{tab:doc_level_split_stats}
\begin{tabular}{lrrrrrrr}
\toprule
\multirow{2}{*}{Dataset}
& \multicolumn{2}{c}{Train}
& \multicolumn{2}{c}{Validation}
& \multicolumn{2}{c}{Test}
& \multirow{2}{*}{Average \# Q / Doc} \\
\cmidrule(lr){2-3}
\cmidrule(lr){4-5}
\cmidrule(lr){6-7}
& \# Docs & \# Q
& \# Docs & \# Q
& \# Docs & \# Q
& \\
\midrule
FetaTab      & 87  & 104 & 87  & 100 & 697   & 812   & 1.17 \\
LongDocURL   & 40  & 209 & 40  & 226 & 316   & 1,890 & 5.87 \\
MMLongBench-Doc  & 13  & 123 & 13  & 103 & 108   & 847   & 8.01 \\
PaperTab     & 31  & 36  & 31  & 37  & 245   & 320   & 1.28 \\
PaperText    & 109 & 292 & 109 & 300 & 869   & 2,212 & 2.58 \\
\bottomrule
\end{tabular}
\end{table*}
\begin{table*}[t]
\centering
\small
\setlength{\tabcolsep}{4pt}
\caption{\textbf{Performance variation across agent configurations.} 
For each agent count, the bold row reports the headline result (single configuration when only one is valid, otherwise the average across valid configurations). 
Indented rows below show the per-configuration breakdown for multi-configuration cases. $\Delta$ denotes the improvement of PE2 + CEPO over the baseline. 
``--'': for 1-agent configuration, we did not apply PE2 or CEPO to compare performance on multi-agent configurations.
Averaged results are also visually plotted in Fig.~\ref{fig:scaling}.}
\label{tab:agent_scaling}
\resizebox{\textwidth}{!}{
\begin{tabular}{c l c c c c}
\toprule
\textbf{\#Agents} & \textbf{Configuration} & \textbf{Baseline} & \textbf{PE2} & \textbf{PE2 + CEPO} & $\Delta$ \\
\midrule
1 & \textbf{\textsc{GENERAL}} & \textbf{54.3} & -- & -- &  \\
\midrule
2 & \textbf{\textsc{GENERAL} + \textsc{SUM}} & \textbf{53.0} & \textbf{55.5} & \cellcolor{cyan!10}\textbf{56.9} & {\color{green!60!black}+\textbf{3.9}} \\
\midrule
\multirow{3}{*}{3}
  & \textbf{Average}                            & \textbf{53.6} & \textbf{53.2} & \cellcolor{cyan!10}\textbf{57.7} & {\color{green!60!black}+\textbf{4.1}} \\
  & \quad \textsc{GENERAL} + \textsc{TEXT} + \textsc{SUM}               & 52.2    & 47.5 & \cellcolor{cyan!10}55.5 & {\color{green!60!black}+3.3} \\
  & \quad \textsc{GENERAL} + \textsc{IMAGE} + \textsc{SUM}              & 55.0    & 58.8 & \cellcolor{cyan!10}59.9 & {\color{green!60!black}+4.9} \\
\midrule
\multirow{4}{*}{4}
  & \textbf{Average}                            & \textbf{54.2} & \textbf{57.5} & \cellcolor{cyan!10}\textbf{58.7} & {\color{green!60!black}+\textbf{4.5}} \\
  & \quad \textsc{GENERAL} + \textsc{TEXT} + \textsc{IMAGE} + \textsc{SUM}       & 55.4    & 57.7 & \cellcolor{cyan!10}59.6 & {\color{green!60!black}+4.2} \\
  & \quad \textsc{GENERAL} + \textsc{CRITIC} + \textsc{TEXT} + \textsc{SUM}      & 52.1    & 56.1 & \cellcolor{cyan!10}57.0 & {\color{green!60!black}+4.9} \\
  & \quad \textsc{GENERAL} + \textsc{CRITIC} + \textsc{IMAGE} + \textsc{SUM}     & 55.1    & 58.7 & \cellcolor{cyan!10}59.6 & {\color{green!60!black}+4.5} \\
\midrule
5 & \textbf{\textsc{GENERAL} + \textsc{CRITIC} + \textsc{TEXT} + \textsc{IMAGE} + \textsc{SUM}} & \textbf{54.1} & \textbf{57.9} & \cellcolor{cyan!10}\textbf{61.0} & {\color{green!60!black}+\textbf{6.9}} \\
\bottomrule
\end{tabular}
}
\end{table*}
\begin{table}[t]
\centering
\caption{\textbf{Results by optimization budget.} Opt: \# optimizer API calls, LLM: \# MDocAgent internal API calls.}
\label{tab:opt_cost}
\vspace{0.5em}
\resizebox{0.5\linewidth}{!}{
\begin{tabular}{c|cc|cc|c}
\toprule
\multirow{2}{*}{Method} 
& \multirow{2}{*}{$T_C$} 
& \multirow{2}{*}{$T_E$}
& \multicolumn{2}{c|}{\# API Calls} 
& \multirow{2}{*}{Acc.} \\
\cmidrule(lr){4-5}
& & & Opt. & LLM & \\
\midrule
\multirow{2}{*}{PE2}
& 0 & 3 & 10 & 5,714 & 57.9 \\
& 0 & 6 & 22 & 11,945 & 57.4 \\
\midrule
\multirow{4}{*}{+ CEPO}
& 1 & 1 & 2 & 3,120 & 60.1 \\
& 2 & 2 & 12 & 7,288 & 60.2 \\
& \cellcolor{cyan!10}3 & \cellcolor{cyan!10}3 & \cellcolor{cyan!10}20 & \cellcolor{cyan!10}11,458 & \cellcolor{cyan!10}61.0 \\
& 4 & 4 & 28 & 15,616 & 58.6 \\
\bottomrule
\end{tabular}
}
\end{table}
\begin{table}[t]
\centering
\caption{\textbf{Inference cost by optimization method.} Throughput (sec/sample): average wall-clock time per sample. Output tokens: average number of output tokens per sample.}
\label{tab:inf_cost}
\vspace{0.5em}
\resizebox{0.55\linewidth}{!}{
\begin{tabular}{l|c|c|c}
\toprule
Method & \makecell{Throughput\\(sec/sample)} & \makecell{Output tokens\\(tok/sample)} & Acc.  \\
\midrule
Baseline & 15.92 & 847.7 & 54.1 \\
\midrule
PE2    & 13.35 & 595.8 & 57.9 \\
\rowcolor{cyan!10}
+ CEPO   & 15.61 & 824.7 & \textbf{61.0} \\
\bottomrule
\end{tabular}
}
\end{table}
\begin{table}[t]
\centering
\caption{\textbf{Analysis on the optimization schedule.} Two design axes are varied: schedule type (sequential vs.\ alternating) and which role is optimized first (coordinator vs.\ executor). All schedules use the same total budget of six iterations.}
\label{tab:schedule_ablation}
\begin{tabular}{l c c}
\toprule
Schedule & Coordinator-first & Executor-first \\
\midrule
Sequential   & \cellcolor{cyan!10}\textbf{61.0} & 60.7 \\
Alternating  & 60.1                 & 58.2 \\
\bottomrule
\end{tabular}
\end{table}
\begin{table}[t]
\centering
\caption{\textbf{Ablation on the target executor set $\mathcal{I}$.}}
\label{tab:executor_subset}
\begin{tabular}{c l c c}
\toprule
$|\mathcal{I}|$ & Target executor set $\mathcal{I}$ & PE2 + CEPO & PE2 \\
\midrule
\multirow{5}{*}{1} 
& \cellcolor{cyan!10}\textsc{GENERAL} (ours) & \cellcolor{cyan!10}\textbf{61.0} & 57.9\\
& \textsc{CRITIC}         & 58.0 & 57.9 \\
& \textsc{TEXT}           & 59.1 & 55.3 \\
& \textsc{IMAGE}          & 58.0 & 54.2 \\
& \textsc{SUM}            & 55.8 & 57.5\\
\midrule
2 & \textsc{GENERAL} $\to$ \textsc{TEXT}                              & 59.7 &56.1 \\
3 & \textsc{GENERAL} $\to$ \textsc{TEXT} $\to$ \textsc{IMAGE}                  & 60.0 & 56.3 \\
4 & \textsc{GENERAL} $\to$ \textsc{CRITIC} $\to$ \textsc{TEXT} $\to$ \textsc{IMAGE}     & 57.5 & 57.6 \\
5 & \textsc{GENERAL} $\to$ \textsc{CRITIC} $\to$ \textsc{TEXT} $\to$ \textsc{IMAGE} $\to$ \textsc{SUM} & 56.9 & 55.3\\
\bottomrule
\end{tabular}
\end{table}
\begin{table}[t]
\centering
\small
\setlength{\tabcolsep}{8pt}
\renewcommand{\arraystretch}{1.1}
\caption{\textbf{Prompt length controlled comparison.} 
\#~Tokens: total tokens of optimized prompts.}
\label{tab:length_controlled}
\begin{tabular}{l|c|c}
\toprule
Method & \# Tokens & Accuracy \\
\midrule
Baseline & 226 & 54.1 \\
\midrule
PE2 (short) & 631 & 57.9 \\
PE2 (long) & 1410 & 54.0 \\
\midrule
\rowcolor{cyan!10}
PE2 + CEPO & 1408 & \textbf{61.0} \\
\bottomrule
\end{tabular}

\end{table}

\section{Further analysis}
\subsection{Cost analysis}
\label{sec:supp_cost_analysis}
\paragraph{Optimization cost comparison.}
Table~\ref{tab:opt_cost} examines accuracy as we vary the optimization budget $(T_C, T_E)$. 
We adopt PE2's default iteration budget for each phase of $(T_C{=}3, T_E{=}3)$ for CEPO, which matches the empirical plateau: pushing the budget further to $(T_C{=}4, T_E{=}4)$ yields no improvement and in fact degrades accuracy from $61.0$ to $58.6$.
When matched against PE2 under comparable budgets, CEPO's gains do not stem from spending more compute. CEPO with $(T_C{=}1, T_E{=}1)$ already improves over PE2 with $(T_C{=}0, T_E{=}3)$ by $2.2$ points ($60.1$ vs.\ $57.9$) while using far fewer API calls, showing that even a minimal optimization budget yields a substantial benefit. At a near-identical optimization cost to PE2 with $(T_C{=}0, T_E{=}6)$, CEPO with $(T_C{=}3, T_E{=}3)$ still leads by $3.6$ points ($61.0$ vs.\ $57.4$), confirming that the advantage comes from more effective optimization rather than from an inflated budget.

\paragraph{Inference cost comparison.}
Table~\ref{tab:inf_cost} reports the inference cost of each method, measured as average wall-clock time per sample. Since input text tokens are dominated by image tokens in the MDocAgent pipeline, throughput is driven primarily by intermediate output length rather than by prompt length. 
CEPO produces intermediate outputs comparable in length to the baseline (PE2 + CEPO: 824.7 vs. baseline: 847.7 tokens) yet improves accuracy by 6.9 points, while vanilla PE2 reaches lower accuracy (57.9) with shorter intermediates (595.8 tokens). 
This indicates that prompts optimized with CEPO elicit richer intermediate reasoning at no additional inference cost compared to the baseline, whereas vanilla PE2's compactness comes at an accuracy cost.

\subsection{Analysis on design choices}
\paragraph{Analysis on optimization schedule.}
\label{sec:supp_optimization_schedule}
Table~\ref{tab:schedule_ablation} ablates the optimization schedule along two axes: schedule type (sequential vs.\ alternating) and starting role (coordinator-first vs.\ executor-first). 
Here, alternating schedule denotes optimizing each agent in turn for a single iteration.
Sequential schedules outperform alternating ones, as letting each role converge before switching avoids the instability of co-adapting two interdependent prompts. 
Optimizing the coordinator first is also preferable, since fixing the high-level pipeline behavior early provides a stable target against which the executors can specialize. Our choice combines both, yielding the highest accuracy of $61.0$.

\paragraph{Analysis on target executor set $\mathcal{I}$.}
\label{sec:supp_optimization_agent_set}
Table~\ref{tab:executor_subset} ablates the choice of the target executor set $\mathcal{I}$, the set of agents whose executor prompts are optimized. For $|\mathcal{I}|=1$, we report results for every agent individually. For $|\mathcal{I}|>1$, agents are added incrementally in descending order of their single-agent accuracy, and their executor prompts are then optimized sequentially along the topological order of the MDocAgent pipeline (\textsc{GENERAL} $\to$ \textsc{CRITIC} $\to$ \textsc{TEXT, IMAGE} $\to$ \textsc{SUM}), with each prompt optimized while all others remain fixed.
Among single-agent targets, optimizing the \textsc{GENERAL} executor yields the highest accuracy ($61.0$), outperforming the next-best target (\textsc{TEXT}, $59.1$) by $1.9$ points and the weakest target (\textsc{SUM}, $55.8$) by $5.2$ points. 
This gap suggests that the \textsc{GENERAL} agent, which produces the initial document-level answer that subsequent agents refine, is the most impactful component to optimize in the pipeline.
Notably, vanilla PE2 exhibits larger performance degradation as more executor prompts are jointly optimized, whereas CEPO mitigates this instability and maintains comparatively robust performance in the multi-agent setting.
Extending the target set beyond \textsc{GENERAL} does not yield further gains, supporting $|\mathcal{I}|=1$ as our default and leaving joint optimization of multiple executor prompts to future work.

\paragraph{CEPO's gains are not from longer prompts.}
A natural concern is whether CEPO's advantage simply comes from feeding longer prompts to the MAS. 
To isolate this effect, we compare three configurations at different prompt length budgets. ``PE2 (short)'' is the default vanilla PE2 setup.
``PE2 (long)'' runs vanilla PE2 with an extended budget so that its optimized prompts are at least as long as CEPO's total prompt length, controlling for the possibility that gains come from longer prompts alone. 
CEPO uses our default $(T_C{=}3, T_E{=}3)$ configuration. 
As shown in Table~\ref{tab:length_controlled}, simply allowing vanilla PE2 to produce longer prompts not only fails to close the gap, but actually \emph{hurts} performance: ``PE2 (long)'' drops to 54.0, falling below even the no-optimization baseline (54.1), suggesting that naively extending prompt length does not necessarily improve performance. 
Meanwhile, additionally applying CEPO results in accuracy of 61.0 with a comparable token budget, demonstrating that the gains stem from \emph{how} prompt parameters are structured and optimized, not from the sheer length of the prompts being optimized.

\begin{table}[t]
\centering
\small
\caption{\textbf{Coordinator prompt transferability.} $p_C^\star$ optimized on MMLongBench-Doc transfers to four other benchmarks with negligible loss compared to dataset-specific optimization.}
\begin{tabular}{lcccc}
\toprule
\textbf{Coordinator Prompt $p_C^\star$} & \textbf{LongDocURL} & \textbf{PaperTab} & \textbf{PaperText} & \textbf{FetaTab} \\
\midrule
Baseline (no coordinator) & 74.9 & 46.3 & 72.1 & 78.8 \\
\midrule
Cross-dataset ($p_C^\star$ from MMLongBench-Doc) & 74.8 & 49.7 & 73.4 & 79.4 \\
In-domain ($p_C^\star$ from target dataset) & \textbf{75.4} & \textbf{53.7} & \textbf{76.4} & \textbf{80.8} \\
\bottomrule
\end{tabular}

\label{tab:coordinator_transfer}
\end{table}
\noindent\textbf{Cross-dataset coordinator transfer.}
Table~\ref{tab:coordinator_transfer} evaluates transfer of a coordinator prompt optimized on MMLongBench-Doc to four other document benchmarks against in-domain optimization. The transferred CEPO configuration achieves an average score of 69.3, compared with 71.6 for in-domain optimization, a gap of 2.3 points. These results provide evidence that the learned coordination instructions can be reused across document benchmarks.

\paragraph{Full optimized prompts.}
We provide the final optimized prompts in \S\ref{sec:final_optimized_mdocagent_prompts}. 
Qualitatively, the coordinator prompt mainly learns \emph{what} agents should coordinate at the team level, while executor prompts primarily learn \emph{how} each agent should focus on during execution.

\section{Prompts}
\label{sec:prompts}

\subsection{Evaluator LLM prompts}
\label{sec:evaluator_llm_prompt}

Table~\ref{tab:evaluator_prompt} provides the input prompt used for the evaluator LLM.

\begin{table}[h]
\centering
\caption{Evaluator LLM prompt.}
\label{tab:evaluator_prompt}
\small
\setlength{\tabcolsep}{6pt}
\renewcommand{\arraystretch}{1.15}
\begin{tabularx}{\linewidth}{>{\bfseries}p{0.18\linewidth} X}
\toprule
Evaluator Prompt &
{\fontfamily{cmtt}\selectfont\fontsize{8pt}{9.5pt}\selectfont
Question: \{question\}

Predicted Answer: \{answer\}

Ground Truth Answer: \{gt\}

\medskip
Please evaluate if the predicted answer is correct compared to the ground truth.

Score the answer on:

Binary correctness (0-1): 1 if the answer is correct, 0 if it is incorrect

\medskip
Return only a string with these scores in a dictionary and can be parsed by json.loads, e.g. \{\{"binary\_correctness": 1\}\}
}
\\
\bottomrule
\end{tabularx}
\end{table}

\newpage
\subsection{Optimizer meta-instruction for optimizing coordinator prompts}
\label{sec:optimizer_llm_prompt}

Tables~\ref{tab:optimizer_prompt} and~\ref{tab:docagent_shared_optimizer_prompt} provide the meta-instructions used for the optimizer LLM for MDocAgent and DocAgent.

\begin{table}[h]
\centering
\caption{\textbf{Meta-instruction for optimizing the coordinator prompt on MDocAgent.}}
\label{tab:optimizer_prompt}
\scriptsize
\setlength{\tabcolsep}{5pt}
\renewcommand{\arraystretch}{1.12}
\begin{tabularx}{\linewidth}{>{\bfseries}p{0.18\linewidth} X}
\toprule
Actor Optimizer Prompt &
{\fontfamily{cmtt}\selectfont
Optimization target: ONE shared multi agent system prompt used by GENERAL, CRITICAL, TEXT, IMAGE, and SUM.

\medskip
\#\#\# Agent System Prompt Optimization Objective

You are optimizing one shared multi-agent system prompt for a multi-agent document QA pipeline.

This shared prompt is prepended to ALL five agents (GENERAL, CRITICAL, TEXT, IMAGE, SUM). Each agent also receives its own agent-specific prompt with role-specific instructions.

The goal is to improve system-level behavior across the trace: uncertainty preservation, modality discipline, evidence handling, handoff, conflict surfacing, and final synthesis.

Do not primarily optimize output format, templates, XML wrappers, field names, or rigid reporting schemas. It is acceptable to add short role-dependent guidance only when it defines division of labor across the system, not detailed role-specific heuristics.

Prefer compact shared rules that help the pipeline make better decisions over formatting instructions that only change how answers look.

\medskip
\#\#\# Brevity as an Optimizer-Only Constraint

Brevity is a constraint on how you edit and select the new shared system prompt. It is not, by itself, a rule that should be inserted into the produced prompt.

Do not add self-referential brevity instructions such as ``be concise,'' ``keep outputs minimal,'' ``use a short format,'' or similar rules unless failure analysis shows that runtime verbosity is itself causing a shared pipeline failure.

Do not mention token budgets, prompt length, minimality, or concision inside the produced shared system prompt unless that is directly justified by the failure pattern.

Instead, satisfy the brevity constraint by deleting, merging, or rewriting redundant shared rules while preserving or improving coordination quality. When two candidate prompts are likely to behave similarly, prefer the shorter one. When a useful change can be made either by revising an existing rule or by adding a new one, prefer revising the existing rule.

Hard limit: the optimized shared system prompt (`new\_prompt`) must remain under 600 tokens.

\medskip
\#\#\# Multimodal Failure Case Format

When failure examples include retrieved images, each case may attach the images once near the top of that case using aliases such as `case1\_image1`, `case1\_image2`, etc. Treat those aliases as the actual images retrieved for that failure case. Later agent input traces may reference the same aliases in `referenced\_images`; do not expect repeated image blocks inside every agent input. Use the case-local aliases to reason about what GENERAL, CRITICAL, and IMAGE could see, while keeping the prompt edit focused on shared multi-agent behavior.

\medskip
\#\#\# Multimodal Optimizer Attachments

The optimizer request includes 20 actual image(s). They are attached to this request in the same order as the image mapping shown inside the failure examples. When analyzing a case, use only the images mapped to that example and only when the target agent input shows those images were runtime-visible.

\medskip
\#\#\# Step 1 (Analyze end-to-end failures)

For each example, inspect the full multi-agent trace and classify the final-answer failure. Decide whether the optimized target prompt could have changed the trace toward the gold answer. Separate target-actionable failures from downstream-only, other-agent-only, and evidence-missing cases.

\medskip
Current prompt:

\{CURRENT PROMPT\}

\medskip
Failure examples:

\{FAIL EXAMPLES INCLUDED\}

\medskip
\#\#\# Step 2 (Produce optimized prompt)

Return ONLY a JSON object with the keys: \{"reasoning": "...", "new\_prompt": "...", "new\_history": "..."\}
}
\\
\bottomrule
\end{tabularx}
\end{table}
\newpage
\begin{table}[h]
\centering
\caption{\textbf{Meta-instruction for optimizing the coordinator prompt on DocAgent.}}
\label{tab:docagent_shared_optimizer_prompt}
\scriptsize
\setlength{\tabcolsep}{5pt}
\renewcommand{\arraystretch}{1.12}
\begin{tabularx}{\linewidth}{>{\bfseries}p{0.18\linewidth} X}
\toprule
Pipeline Optimizer Prompt &
{\fontfamily{cmtt}\selectfont
Optimization target: one shared DocAgent pipeline instruction. It is composed into the actual system message with the original DocAgent system prompt.

\medskip
\#\#\# Agent System Prompt Optimization Objective

You are optimizing one shared multi-agent system prompt for a multi-agent document QA pipeline.

This shared prompt is prepended to ALL three agents (actor, reviewer, reflector). Each agent also receives its own agent-specific prompt with role-specific instructions.

The goal is to improve system-level behavior across the trace: evidence handling, handoff, conflict surfacing, and final synthesis.

Encourage each agent to act as evidence support for the final synthesis. Non-final agents should preserve information that may help the final answerer, including verified evidence, partial support, and conflicts.

Optimize the shared prompt toward better evidence preservation. The resulting prompt should help the final synthesizer recover the strongest answer supported by the available context.

Do not primarily optimize output format, templates, XML wrappers, field names, or rigid reporting schemas.

It is acceptable to add short role-dependent guidance only when it defines division of labor across the system, not detailed role-specific heuristics.

Prefer compact shared rules that help the pipeline make better decisions over formatting instructions that only change how answers look.

\medskip
\#\#\# Brevity as an Optimizer-Only Constraint

Brevity is a constraint on how you edit and select the new shared system prompt.

It is not, by itself, a rule that should be inserted into the produced prompt.

Do not add self-referential brevity instructions such as ``be concise,'' ``keep outputs minimal,'' ``use a short format,'' or similar rules unless failure analysis shows that runtime verbosity is itself causing a shared pipeline failure.

Do not mention token budgets, prompt length, minimality, or concision inside the produced shared system prompt unless that is directly justified by the failure pattern.

Instead, satisfy the brevity constraint by deleting, merging, or rewriting redundant shared rules while preserving or improving coordination quality.

When two candidate prompts are likely to behave similarly, prefer the shorter one.

When a useful change can be made either by revising an existing rule or by adding a new one, prefer revising the existing rule.

Hard limit: the optimized shared system prompt (`new\_prompt`) must remain under 600 tokens.

\medskip
\#\#\# Step 1 (Analyze end-to-end failures)

For each example, inspect the actor/reviewer/reflection trace and classify the final-answer failure. Decide whether shared pipeline behavior could have changed the trace toward the gold answer. Separate prompt-actionable failures from evidence-missing or judge-only issues.

The optimized prompt must preserve these template placeholders: \{agent\_role\} and \{agent\_specific\_prompt\}.

\medskip
Current prompt:

\{current\_prompt\}

\medskip
Failure examples:

\{failure\_examples\}

\medskip
\#\#\# Step 2 (Produce optimized prompt)

\{history\_section\}

\medskip
Return ONLY a JSON object with keys ``reasoning'', ``new\_prompt'', and ``new\_history''.
}
\\
\bottomrule
\end{tabularx}
\end{table}

\newpage
\subsection{Optimizer meta-instruction for optimizing executor prompts}
\label{sec:agent_optimizer_llm_prompt}
Table~\ref{tab:agent_optimizer_prompt} provides the meta-instruction used for optimizing agent-specific executor prompts.

\begin{table}[h]
\centering
\caption{\textbf{Meta-insturction for optimizing the executor prompt on MDocAgent.}}
\label{tab:agent_optimizer_prompt}
\scriptsize
\setlength{\tabcolsep}{5pt}
\renewcommand{\arraystretch}{1.12}
\begin{tabularx}{\linewidth}{>{\bfseries}p{0.18\linewidth} X}
\toprule
Pipeline Optimizer Prompt &
{\fontfamily{cmtt}\selectfont
Target agent: \{TARGET AGENT\}

\medskip
You are optimizing one target agent prompt in a multi-agent system.

Your task is to optimize only the target agent's executor prompt: the how-to instructions that tell this specific agent how to perform its assigned subtask under the given context.

Analyze the failure examples using only the target-agent input and target-agent output shown in each case. Do not rely on hidden upstream reasoning, downstream corrections, missing pages, unavailable images, or information that was not visible to the target agent. Infer the execution behavior that would have prevented each failure, then generalize those behaviors into reusable executor-level instructions.

\medskip
Target optimized prompt granularity:

- The final prompt should be concrete and operational, not merely a generic multimodal QA prompt.

- It should provide enough task-specific guidance for the target agent to avoid recurring failure patterns.

- Merge similar behaviors into broader rules when possible, but keep the resulting rules specific enough to guide execution.

- Include procedural details when they clarify how the agent should select evidence, reason over multimodal inputs, and produce its answer.

- Avoid overfitting to individual examples or encoding narrow case-specific fixes.

- Avoid making the prompt unnecessarily long, but do not omit useful execution guidance just to keep it minimal.

\medskip
Prompt organization:

- Write the optimized prompt as an executor-facing how-to guide.

- Prefer a compact structure with:

\quad 1. a brief statement of the target agent's task;

\quad 2. a ``Core behavior'' section describing the main execution rules the agent should follow;

\quad 3. a ``Procedure'' section describing the order in which the agent should apply those rules.

- The ``Core behavior'' section should contain stable behavioral principles derived from the failures.

\medskip
Keep the structure focused on how this target agent should execute its own task.

The optimized prompt should stay at the executor level. It should describe how the target agent should perform its own multimodal document QA task under the given context. Do not redefine the overall multi-agent system, assign roles to other agents, change global routing, or add unsupported capabilities.

\medskip
\#\#\# Step 1 (Analyze target-agent failures)

For each example, inspect only the target agent input/output shown in the failure case. Decide what prompt behavior would have made that target agent answer correctly from the same input. Do not rely on hidden upstream or downstream agent traces.

\medskip
Current prompt: \{CURRENT PROMPT\}

\medskip
Failure examples:

\{FAIL EXAMPLES INCLUDED\}

\medskip
\#\#\# Step 2 (Produce optimized prompt)

Return ONLY a JSON object with the keys: \{"reasoning": "...", "new\_prompt": "...", "new\_history": "..."\}
}
\\
\bottomrule
\end{tabularx}
\end{table}

\newpage
\subsection{LLM-generated initial coordinator prompts}
\label{sec:llm_gen_coord_prompt}

Tables~\ref{tab:seed_coordinator_prompt} and~\ref{tab:docagent_shared_seed_prompt} show the seed coordinator prompts for MDocAgent and DocAgent, generated by GPT-5-mini and used as the initial coordinator prompt $p^{(0)}_C$ for optimization.

\begin{table}[h]
\centering
\caption{\textbf{Initial coordinator prompt $p^{(0)}_C$ for MDocAgent.}}
\label{tab:seed_coordinator_prompt}
\small
\setlength{\tabcolsep}{6pt}
\renewcommand{\arraystretch}{1.15}
\begin{tabularx}{\linewidth}{>{\bfseries}p{0.18\linewidth} X}
\toprule
MDocAgent Coordinator Prompt &
{\fontfamily{cmtt}\selectfont\fontsize{8pt}{9.5pt}\selectfont
Current prompt:

You are one agent in a multi-agent document question answering system.

\medskip
There are five agents:

- GENERAL: can broadly inspect both text and image evidence

- CRITICAL: can broadly inspect both text and image evidence

- TEXT: can only inspect text evidence extracted from the document image, including OCR text

- IMAGE: can only inspect the relevant document image

- SUM: final judge and answer composer; can only view the discussion history from GENERAL, TEXT, and IMAGE

\medskip
You are the \{agent\_role\} agent.

\medskip
\{agent\_specific\_prompt\}
}
\\
\bottomrule
\end{tabularx}
\end{table}
\begin{table}[h]
\centering
\caption{\textbf{Initial coordinator prompt $p^{(0)}_C$ for DocAgent.}}
\label{tab:docagent_shared_seed_prompt}
\small
\setlength{\tabcolsep}{6pt}
\renewcommand{\arraystretch}{1.15}
\begin{tabularx}{\linewidth}{>{\bfseries}p{0.18\linewidth} X}
\toprule
DocAgent Coordinator Prompt &
{\fontfamily{cmtt}\selectfont\fontsize{8pt}{9.5pt}\selectfont
You are part of DocAgent, a document question answering pipeline for long multimodal documents.

\medskip
The pipeline has three prompt-driven stages:

- ACTOR: studies the document outline, uses search, section, page-image, image, and table-image tools to gather evidence, and produces a concise answer inside \textless final\_result\textgreater.

- REVIEWER: checks the actor answer against document evidence using only necessary tools, then either confirms or corrects the final answer inside \textless final\_result\textgreater.

- REFLECTION: when the reviewer changes the actor answer, updates a concise reusable guideline that may help future questions.

\medskip
You are the \{agent\_role\}.

\medskip
Follow the core system instruction below while preserving the shared pipeline behavior:

\medskip
\{agent\_specific\_prompt\}
}
\\
\bottomrule
\end{tabularx}
\end{table}

\newpage
\subsection{{Qualitative Examples of Optimized Prompts}}
\label{sec:final_optimized_mdocagent_prompts}

Tables~\ref{tab:mdocagent_final_coordinator_prompt} and~\ref{tab:mdocagent_final_executor_prompt} show the final optimized coordinator and executor prompts for MDocAgent.

\begin{xltabular}{\linewidth}{
>{\fontfamily{cmtt}\selectfont\fontsize{8pt}{9.5pt}\selectfont\raggedright\arraybackslash}X
}
\caption{\textbf{Optimized coordinator prompt for MDocAgent.}}
\label{tab:mdocagent_final_coordinator_prompt}
\\
\toprule
\endfirsthead

\toprule
\multicolumn{1}{l}{\fontsize{8pt}{9.5pt}\selectfont\itshape Continued from previous page.} \\
\midrule
\endhead

\bottomrule
\endfoot

You are one agent in a multi-agent document question answering system.
\\

There are five agents:
\\

- GENERAL: can broadly inspect both text and image evidence
\\

- CRITICAL: can broadly inspect both text and image evidence
\\

- TEXT: can only inspect text evidence extracted from the document image, including OCR text
\\

- IMAGE: can only inspect the relevant document image
\\

- SUM: final judge and answer composer; can only view the discussion history from GENERAL, TEXT, and IMAGE
\\

You are the \{agent\_role\} agent.
\\

Shared rules for all agents:
\\

- Use only the evidence available to your role. Do not claim to have seen text, images, pages, or prior discussion that your role cannot access.
\\

- Separate direct evidence from inference. If a value, phrase, identifier, linkage, or conclusion is not explicitly supported by available evidence, do not present it as found.
\\

- Preserve uncertainty. If evidence is missing, ambiguous, partially visible, or insufficient for the requested comparison, ranking, count, sum, ratio, or citation lookup, say so and prefer not answerable over guessing.
\\

- Keep scope local to the question. Anchor answers to the most relevant page, section, table, figure, footnote, or image region, and do not replace that local scope with broader document-level statistics or other pages unless the evidence clearly requires it.
\\

- Do not link items across different pages, sections, tables, or map regions unless the evidence explicitly connects them. If multiple local candidates exist, report the ambiguity instead of merging them into one answer.
\\

- When a quoted statement, marker, footnote, label, or pointer appears, resolve it using the nearest explicit local anchor tied to that statement; do not substitute a nearby earlier citation just because it is topically related.
\\

- If you cite numbers, names, IDs, case numbers, footnotes, or quoted text, copy them exactly from the evidence when possible. Do not normalize, reformat, convert units, or ``clean up'' identifiers unless explicitly asked.
\\

- For computations or comparisons, use only explicitly supported operands and units. If a needed operand is missing, obscured, incomplete, or on a different unit basis, state that the result cannot be determined reliably.
\\

- Be explicit about coverage limits: note if only some pages/images are visible, or if the answer would require evidence not retrieved.
\\

- When text and image evidence agree, use that agreement. When they conflict, or when another agent's likely claim is unsupported by your evidence or mismatched to the local anchor/scope, explicitly flag the conflict or lack of support.
\\

- If your role cannot fully answer the question, provide the most relevant grounded observations and identify what remains unresolved for other agents or final synthesis.
\\

- For SUM: synthesize only from the discussion history. Prefer claims that are grounded, locally scoped, and corroborated. Do not resolve conflicts by choosing the most specific unsupported claim, and do not turn partial evidence into a definite numeric or citation answer.
\\

\{agent\_specific\_prompt\}
\\

\end{xltabular}
\newpage
\begin{xltabular}{\linewidth}{
>{\fontfamily{cmtt}\selectfont\fontsize{8pt}{9.5pt}\selectfont\raggedright\arraybackslash}X
}
\caption{\textbf{Optimized executor prompt for the GENERAL agent in MDocAgent.}}
\label{tab:mdocagent_final_executor_prompt}
\\
\toprule
\endfirsthead

\toprule
\multicolumn{1}{l}{\fontsize{8pt}{9.5pt}\selectfont\itshape Continued from previous page.} \\
\midrule
\endhead

\bottomrule
\endfoot

You are the GENERAL agent in a multimodal document QA system. You can inspect both OCR/text and the visible document images. Answer only from the evidence actually visible to you.
\\

Core behavior:
\\

1. Ground every answer in the most local relevant evidence: the exact visible page, slide, figure, table, header, footer, caption, panel, or image region tied to the question.
\\

2. Treat OCR text and the image as joint evidence. If a page number, label, count cue, identifier, row label, or year appears in either source, use it.
\\

3. Identify the exact target scope before answering: a specific page, last page, row, subject, figure, chapter, year, page range, component, or subset named in the question.
\\

4. Do not replace the asked scope with a broader or different one. In particular, do not substitute whole-document totals, benchmark totals, or unrelated page content unless the question explicitly asks for them.
\\

5. If document coverage is incomplete and the question depends on unseen pages/images or on a complete set, for example: ``all samples,'' ``each chapter,'' ``entire document,'' ``last page,'' answer: Not answerable from the visible evidence.
\\

6. Example/case-study figures are not automatically the population of interest. If the question asks about all examples in a subject/category, use the visible index/table/list that enumerates that set, not a single showcased example page.
\\

7. For identifiers, names, quoted text, figure numbers, service specifications, case numbers, and labels, copy exactly as shown. Do not normalize punctuation, spacing, commas, hyphens, abbreviations, or formatting unless explicitly asked.
\\

8. For table questions, first filter to rows whose key field exactly matches the requested entity, subject/component/year/page/chapter/etc. Then extract only the requested column(s) from those rows. Do not mix in values from neighboring rows or other components.
\\

9. For conditional lookups, bind the condition to the exact local anchor. Example: if a question asks ``when X reached Y,'' first find the year/row where Y occurs, then read the corresponding value for that same year/row from the relevant chart/table. Do not switch to a more visually salient nearby number.
\\

10. For percentages/ratios/counts, use only operands explicitly supported by the relevant local evidence. Do not invent denominators. Do not use whole-document totals unless the question explicitly asks for a percentage out of the whole document/benchmark.
\\

11. When counting ``how many distinct examples/samples,'' count unique matching indexed items in the relevant local scope, deduplicating repeated renderings and ignoring unrelated pages.
\\

12. If the question asks about ``all'' or ``each,'' verify that all requested members are visible. If only some members are shown, answer Not answerable from the visible evidence rather than generalizing from the visible subset.
\\

13. If multiple visible candidates exist and the question does not clearly specify which one, state the ambiguity briefly instead of guessing.
\\

14. If evidence is insufficient, ambiguous, truncated, obscured, or requires linking beyond what is visible, say Not answerable from the visible evidence.
\\

15. Prefer a short direct answer. Do not include long reasoning unless needed to explain unanswerability or a brief calculation.
\\

16. Ensure the final answer matches your own grounded evidence. Do not present exploratory or conflicting reasoning that points to a different final answer.
\\

Procedure:
\\

- First identify the exact target scope named in the question.
\\

- Verify that the scope is visible and complete enough.
\\

- Find the nearest governing anchor: exact row label, year, chapter, subject, figure index, or component name.
\\

- If extracting from a table/list, filter rows by exact match before collecting values.
\\

- If computing, use only the visible operands tied to that same local scope.
\\

- If a required scope is not fully visible, answer: Not answerable from the visible evidence.
\\

- Return just the answer, or the answer plus a very short grounding phrase.
\\

\end{xltabular}

\newpage
\subsection{Qualitative comparison of prompts optimized with different objectives.}
Table~\ref{tab:role_aligned_objective_excerpts} provides qualitative evidence for the role-aligned objectives principle (Table~\ref{tab:swap}) by showing how each prompt layer evolves under different optimization objectives. 
Under the \emph{what-to} objective (top row), both the coordinator and executor prompts converge to high-level signals that define desirable system behavior, such as ``use only the evidence available to your role'' or ``preserve the exact target of the question.'' 
In contrast, under the \emph{how-to} objective (bottom row), both layers converge to procedural rules and step-by-step instructions tailored to specific execution patterns. 
The aligned configuration ($p_C$=\emph{what-to}, $p_E$=\emph{how-to}) thus assigns each layer the kind of signal it is structurally suited to carry: high-level coordination at the coordinator and concrete execution at the executor.
\label{sec:role_aligned_objective_excerpts}
\begin{table*}[t]
\centering
\caption{\textbf{Coordinator and executor prompts optimization results under different objectives.}
Rows denote objective types (\emph{what-to} vs. \emph{how-to}), and columns denote prompt layers (coordinator vs. executor).
Blue highlights indicate high-level \whathl{\emph{what-to}} signals defining desirable behavior, and yellow highlights \howhl{\emph{how-to}} signals describing procedural details and rules. \protect\ourshl{Green-shaded cells} mark the role-aligned configuration used in our method.}
\label{tab:role_aligned_objective_excerpts}
\scriptsize
\setlength{\tabcolsep}{3.3pt}
\renewcommand{\arraystretch}{1.12}
\begin{tabularx}{\textwidth}{
@{}>{\raggedright\arraybackslash}p{0.105\textwidth}
>{\raggedright\arraybackslash}X
>{\raggedright\arraybackslash}X@{}}
\toprule
&
\multicolumn{1}{c}{\textbf{Coordinator prompt}} &
\multicolumn{1}{c}{\textbf{Executor prompt}} \\
\midrule

\whattag{WHAT-TO}
&
\cellcolor{oursbg}\promptcell{
You are one agent in a multi-agent document question answering system. \promptdots\ 
- Use only the \whathl{evidence available to your role}; do not claim unseen text, images, pages, or prior discussion.\par
- \whathl{Separate direct evidence} from inference; unsupported values, phrases, identifiers, or conclusions should not be presented as found.\par
- \whathl{Preserve uncertainty}; if evidence is missing, ambiguous, or insufficient, prefer not answerable over guessing.\par
- Keep \whathl{scope local} to the question; do not replace local evidence with broader document-level statistics or unrelated pages.\par
- Do not link items across pages, sections, tables, or map regions unless the evidence explicitly connects them. \promptdots\ 
- If your role cannot fully answer, provide \whathl{grounded observations} and identify \whathl{what remains unresolved} for other agents or final synthesis.\promptdots
}
&
\promptcell{
Answer the user's question using only document content that is visibly established in the provided text and images. \promptdots\ 
Your answer must \whathl{preserve the exact target} of the question: the requested entity type, group, field, location relation, count, title, ratio, or other attribute.\par
\whathl{Favor the most specific evidence} that directly matches the ask.\par
Do not substitute a nearby heading, broader list, related item, parent category, child category, or different semantic type just because it appears relevant. \promptdots\ 
Reflect the document faithfully: when the answer depends on labels, captions, map placement, section membership, figure association, or category wording, use only what is \whathl{directly shown and clearly linked}. \promptdots\ 
If the exact requested content is not visibly established, or if the visible evidence only supports a related but different answer, \whathl{say so rather than guessing}.
}
\\

\midrule
\howtag{HOW-TO}
&
\promptcell{
You are one agent in a multi-agent document question answering system. \promptdots\ 
Before answering, \howhl{check all of the following}: 
1) Do I have the right evidence source for this question?
2) Is the referenced page, region, table, chart, figure, row, item, label, or field \howhl{actually visible} or explicitly quoted?
3) Does the evidence support the \howhl{exact answer requested}, rather than a nearby, broader, parent-category, child-category, or implied one?
4) Am I using the correct \howhl{unit, denominator, item identity}, counting rule, and formatting rule? \promptdots\ 
Do not assume the full document is available; retrieved items may be \howhl{partial, cropped, non-consecutive}, or from different parts of the document.
If the question targets a specific page, region, figure, row, item, or file, verify that the target content is \howhl{visibly present and clearly linked}.
For exact-string tasks, \howhl{copy carefully} and preserve unclear punctuation, separators, spacing, decimals, or digit grouping.
For visual-region questions, \howhl{anchor your reasoning} to the specified region; if it is cropped, occluded, or absent, say so.
If multiple plausible answers exist, \howhl{keep them as candidates} and state what evidence remains unresolved.
For maps, \howhl{first identify} the relevant visible map page or region before naming a place or sight.
For tables, \howhl{identify the table title}, row, and column headers before extracting any cell.
For charts, \howhl{identify title, axes, units}, legend, and series before reading values.
For counts, ratios, or percentages, \howhl{define the unit} before calculating and verify the numerator and denominator. \promptdots\ 
If text and image disagree, \howhl{state the conflict} rather than forcing a single answer.
Use exactly: \howhl{Not answerable} from the provided evidence when the required support is missing or unresolved. \promptdots
}
&
\cellcolor{oursbg}\promptcell{
You can inspect both OCR/text and visible document images. Answer only from evidence actually visible to you. \promptdots\ 
1. \howhl{Ground every answer} in the most local relevant evidence: page, slide, figure, table, header, footer, caption, panel, or image region.
2. Treat OCR text and the image as \howhl{joint evidence}; use page numbers, labels, count cues, identifiers, row labels, or years from either source.
3. \howhl{Identify the exact target scope} before answering: page, last page, row, subject, figure, chapter, year, page range, component, or subset. \promptdots\ 
\howhl{use the visible index/table/list} when the question asks about all examples. \promptdots\ 
7. For exact strings and identifiers, \howhl{copy exactly}; do not normalize punctuation, spacing, hyphens, abbreviations, or formatting.
8. For table questions, \howhl{first filter to rows} whose key field exactly matches the requested entity; then \howhl{extract only} the requested column(s).
9. For conditional lookups, \howhl{bind the condition} to the exact local anchor.
10. For percentages, ratios, and counts, use only visible operands tied to the same local evidence.
11. For distinct examples or samples, \howhl{count unique matching items} and deduplicate repeated renderings.
12. If the question asks about all or each, \howhl{verify that all requested members are visible}; otherwise answer \howhl{Not answerable} rather than generalizing.
13. If multiple visible candidates remain, \howhl{state the ambiguity} instead of guessing. \promptdots\ 
\howhl{Procedure}: first identify the exact target scope; verify that the scope is visible and complete enough; find the \howhl{nearest governing anchor}; extract using the correct structure.
If extracting from a table/list, \howhl{filter rows by exact match} before collecting values.
If a required scope is not fully visible, answer: \howhl{Not answerable} from the visible evidence. \promptdots
}
\\

\bottomrule
\end{tabularx}
\end{table*}

\end{document}